\documentclass[sigconf, nonacm]{acmart}
\usepackage{comment}
\newcommand\vldbdoi{XX.XX/XXX.XX}
\newcommand\vldbpages{XXX-XXX}
\newcommand\vldbvolume{14}
\newcommand\vldbissue{1}
\newcommand\vldbyear{2020}
\newcommand\vldbauthors{\authors}
\newcommand\vldbtitle{\shorttitle} 

\newcommand\vldbpagestyle{plain} 

\usepackage{booktabs} \usepackage{multibib}
\usepackage{xcolor}
\usepackage{booktabs}
\usepackage{adjustbox}
\usepackage{varwidth}
\DeclareCaptionLabelFormat{andtable}{#1~#2  \&  \tablename~\thetable}
\newcites{latex}{Additional References}

\usepackage{mdwlist}
\usepackage{stfloats}

\usepackage{enumitem}

\usepackage{footmisc}
\usepackage{comment}
\usepackage{tabularx}
\usepackage{mathrsfs}
\usepackage{amsmath}
\usepackage{amsfonts}
\usepackage{algorithm}
\usepackage{footmisc}
\usepackage{tikz}
\usepackage{xspace}
\usepackage{graphicx}
\usepackage{multirow}
\usepackage[font={small,bf},labelfont=bf,format=plain]{caption}
\usepackage{subcaption}
\usepackage{pifont}\usetikzlibrary{matrix,decorations.pathreplacing,calc}
\usepackage{mathtools}

\pgfkeys{tikz/mymatrixenv/.style={decoration=brace,every left delimiter/.style={xshift=3pt},every right delimiter/.style={xshift=-3pt}}}
\pgfkeys{tikz/mymatrix/.style={matrix of math nodes,left delimiter=[,right delimiter={]},inner sep=2pt,column sep=1em,row sep=0.5em,nodes={inner sep=0pt}}}
\pgfkeys{tikz/mymatrixbrace/.style={decorate,thick}}

\usepackage{xcolor}
\definecolor{thedarkblue}{RGB}{0,0,120} \definecolor{mydarkblue}{rgb}{0,0.08,0.45} 
\usepackage{hyperref}
\hypersetup{    colorlinks=true,
    linkcolor=black,
    citecolor=black,
    filecolor=mydarkblue,
    urlcolor=mydarkblue}
    
\usepackage{microtype}
\usepackage{balance}

\newcolumntype{L}[1]{>{\raggedright\let\newline\\\arraybackslash\hspace{0pt}}m{#1}}
\newcolumntype{C}[1]{>{\centering\let\newline\\\arraybackslash\hspace{0pt}}m{#1}}
\newcolumntype{R}[1]{>{\raggedleft\let\newline\\\arraybackslash\hspace{0pt}}m{#1}}

\newcommand{\eat}[1]{\ignorespaces}

\newcommand{\dijin}[1]{{\textcolor{black}{#1}}}

\providecommand{\mat}[1]{\boldsymbol{\mathrm{#1}}}\renewcommand{\vec}[1]{\boldsymbol{\mathrm{#1}}}
\providecommand{\sca}[1]{{\mathrm{#1}}}

\DeclareMathOperator*{\minimize}{minimize}
\DeclareMathOperator*{\maximize}{maximize}

\providecommand{\subjectto}{\ensuremath{\text{subject to}}}
\providecommand{\MINof}[1][]{{\displaystyle \minimize_{#1}}}

\providecommand{\MAXof}[1][]{{\displaystyle \maximize_{#1}}}

\DeclareMathOperator{\E}{E}
\DeclareMathOperator{\hugeE}{\mbox{\huge\raise-0.3ex\hbox{E}}}
\DeclareMathOperator{\p}{\mathbb{P}}
\DeclareMathOperator{\hugep}{\mbox{\huge\raise-0.3ex\hbox{$\p$}}}

\DeclareMathOperator{\Std}{Std}

\DeclareMathOperator{\bigO}{O}

\newcommand{\method}{\textsc{AdaMEL}\xspace}

\newcommand{\methodA}{\textsc{AdaMEL-zero}\xspace}
\newcommand{\methodB}{\textsc{AdaMEL-few}\xspace}
\newcommand{\methodC}{\textsc{AdaMEL-hyb}\xspace}

\usepackage{setspace}
\graphicspath{{./}{./graphics/}}
\newcolumntype{H}{>{\setbox0=\hbox\bgroup}c<{\egroup}@{}}

\newcommand{\eg}{\emph{e.g.}}
\newcommand{\ie}{\emph{i.e.}}

\usepackage{tikz}
\usepackage{verbatim}
\usetikzlibrary{arrows}
\usetikzlibrary{shapes,snakes}
\usetikzlibrary{decorations.pathmorphing} \usetikzlibrary{fit}					\usetikzlibrary{backgrounds}	
\usepackage{algorithm}
\usepackage{algorithmicx}
\usepackage{algpseudocode}

\algblockdefx[parallel]{ParFor}{EndPar}[1][]{$\textbf{parallel for}$ #1 $\textbf{do}$}{$\textbf{end parallel}$}
\algrenewcommand{\alglinenumber}[1]{\fontsize{6.5}{7}\selectfont#1}
\algtext*{EndPar}

\algblockdefx[parallel]{parfor}{endpar}[1][]{$\textbf{parallel for}$ #1 $\textbf{do}$}{$\textbf{end parallel}$}
\algrenewcommand{\alglinenumber}[1]{\scriptsize#1:}

\algblockdefx[foreach1]{foreach}{endforeach}[1][]{$\textbf{for}$ #1 $\textbf{do}$}{$\textbf{end for}$}
\algrenewcommand{\alglinenumber}[1]{\scriptsize#1:}

\algnotext{EndFor}
\algnotext{EndIf}
\algnotext{EndProcedure}



\algblockdefx[fornew1]{foreach}{endforeach}
[1][]{\textbf{for} #1}
{\textbf{end for}}

\newtheorem{problem}{Problem}

\begin{abstract}
    Multi-source entity linkage focuses on integrating knowledge from multiple sources by linking the records that represent the same real world entity. This is critical in high-impact applications such as data cleaning and user stitching.
The state-of-the-art entity linkage pipelines mainly depend on supervised learning that requires abundant amounts of training data.
However, collecting well-labeled training data becomes expensive when the data from many sources arrives incrementally over time.  
Moreover, the trained models can easily overfit to specific data sources, and thus fail to generalize to new sources due to significant differences in data and label distributions.
To address these challenges, we present \method, a deep transfer learning framework that learns generic high-level knowledge to perform multi-source entity linkage.
\method models the attribute importance that is used to match entities through an attribute-level self-attention mechanism, and leverages the massive unlabeled data from new data sources through domain adaptation to make it generic and data-source agnostic.
In addition, \method is capable of incorporating an additional set of labeled data to more accurately integrate data sources with different attribute importance.
Extensive experiments show that our framework achieves state-of-the-art results with $8.21\%$ improvement on average over methods based on supervised learning. Besides, it is more stable in handling different sets of data sources in less runtime.

    \label{sec:abstract}
\end{abstract}

\begin{document}

\title{Deep Transfer Learning for Multi-source Entity Linkage via Domain Adaptation}

\author{Di Jin}
\affiliation{  \institution{University of Michigan}
  \city{Ann Arbor}
  \country{USA}
}
\email{dijin@umich.edu}

\author{Bunyamin Sisman}
\affiliation{  \institution{Amazon}
  \city{Seattle}
  \country{USA}
}
\email{bunyamis@amazon.com}

\author{Hao Wei}
\affiliation{  \institution{Amazon}
  \city{Seattle}
  \country{USA}
}
\email{wehao@amazon.com}

\author{Xin Luna Dong}
\affiliation{  \institution{Amazon}
  \city{Seattle}
  \country{USA}
}
\email{lunadong@amazon.com}

\author{Danai Koutra}
\affiliation{  \institution{University of Michigan}
  \city{Ann Arbor}
  \country{USA}
}
\email{dkoutra@umich.edu}

\date{}

\maketitle

\pagestyle{\vldbpagestyle}
\begingroup\small\noindent\raggedright\textbf{PVLDB Reference Format:}\\
\vldbauthors. \vldbtitle. PVLDB, \vldbvolume(\vldbissue): \vldbpages, \vldbyear.\\
\href{https://doi.org/\vldbdoi}{doi:\vldbdoi}
\endgroup
\begingroup
\renewcommand\thefootnote{}\footnote{\noindent
This work is licensed under the Creative Commons BY-NC-ND 4.0 International License. Visit \url{https://creativecommons.org/licenses/by-nc-nd/4.0/} to view a copy of this license. For any use beyond those covered by this license, obtain permission by emailing \href{mailto:info@vldb.org}{info@vldb.org}. Copyright is held by the owner/author(s). Publication rights licensed to the VLDB Endowment. \\
\raggedright Proceedings of the VLDB Endowment, Vol. \vldbvolume, No. \vldbissue\ ISSN 2150-8097. \\
\href{https://doi.org/\vldbdoi}{doi:\vldbdoi} \\
}\addtocounter{footnote}{-1}\endgroup

\section{Introduction}
\label{sec:intro}

Entity linkage (EL), also known as entity resolution, record linkage, entity matching, is a fundamental task in data mining, database, and knowledge integration with numerous applications, including deduplication, data cleaning, user stitching, and more.
The key idea is to identify records across different data sources (\eg, databases, websites, knowledge base, etc.) that represent the \textit{same} real-world entity.
For example, some music websites record the song "Hey Jude" by Paul McCartney with the name abbreviation (i.e., ``P.M.'') while others with the band name (i.e., ``The Beatles'').
As newly-generated data surge over time, accurately consolidating the same entities across semi-structured web sources becomes increasingly important, especially in areas such as knowledge base establishment~\cite{dong2009data,getoor2012entity} and  personalization~\cite{jin2019node2bits}.

\begin{figure}[tp]
\vspace{-.1cm}
\centering
\includegraphics[width=1.\linewidth]{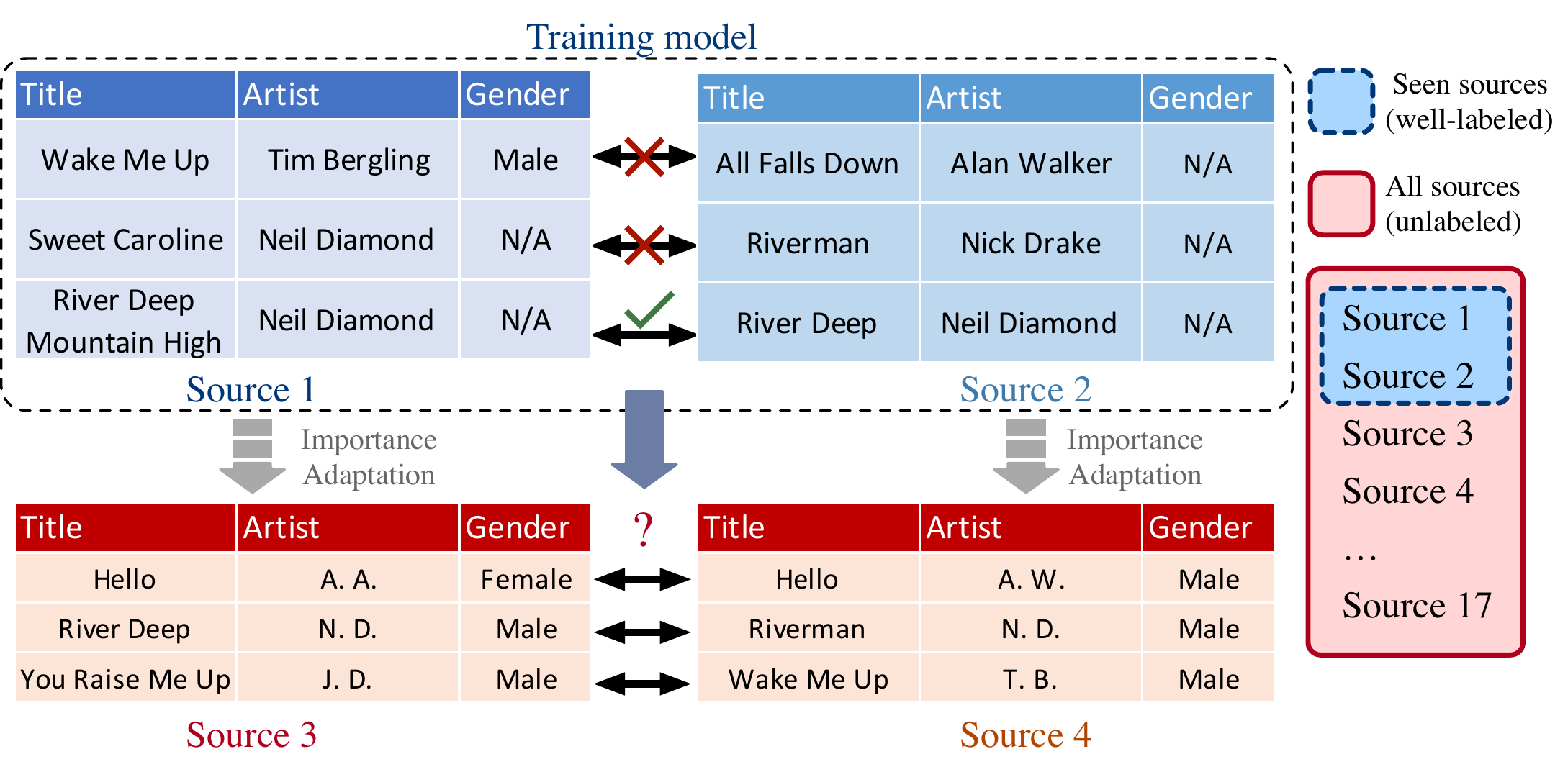}
\vspace{-.6cm}
\caption{
Well-labeled data sources (\eg, blue tables) are generally outnumbered by massive unlabeled data in real-world knowledge integration scenarios. Entity linkage models trained only on well-labeled samples fail to handle new sources with different contexts or formats (\eg, red tables).
Our proposed framework, \method automatically learns the attribute importance that adapts to the massive unlabeled data from different sources during training, and then uses it as the transferable knowledge to perform matching.
}
\label{fig:highlight}
\vspace{-.5cm}
\end{figure}

Methods for solving the entity linkage problem across data sources include rule reasoning~\cite{fan2009reasoning,singh2017synthesizing}, computation of similarity between attributes or schemas~\cite{bilenko2003adaptive}, and active learning~\cite{qian2017active}. 
In particular, recent deep learning approaches that are based on heterogeneous schema matching or word matching~\cite{mudgal2018deep,nie2019deep,li2020deep} have been widely studied. Their promising performance mainly comes from the sophisticated word-level operations such as RNN and Attention~\cite{mudgal2018deep,fu2020hierarchical} to represent token sequences under attributes as the summarization, or the usage of pretrained language models~\cite{kenton2019bert} to better learn the word semantics.
However, all the above learning approaches implicitly assume that the ``matching/non-matching'' info for training records is available (\eg, the music records in source 1 and source 2 shown in the two blue tables of Fig.~\ref{fig:highlight}) and can be queried through the learning process, which does not always hold in practice. In real-world knowledge integration scenarios, new data come incrementally, and it can be either well-labeled (\eg, through manual confirmation) or unlabeled. 
While the existing frameworks can handle high-quality labeled data, they cannot deal with the massive volume of unlabeled and previously unseen data or missing values. As the example shown in Fig.~\ref{fig:highlight}, a model trained on the high-quality labeled data (blue tables) would fail to generalize to the new data sources (red tables) with missing and different attribute values (\ie, ``Artist''), as well as new attributes or attributes that are rarely seen (\ie, ``Gender'').

Motivated by real-world knowledge integration settings,
we consider three key challenges in the multi-data source scenario:
(\textbf{C1}) missing attribute values from unseen data sources; 
(\textbf{C2}) new attributes from unseen data sources; and 
(\textbf{C3}) different value distribution in unseen data sources. Based on these challenges, we seek to tackle the following \textbf{MEL (multi-source entity linkage) problem}: Given labeled data from a limited set of sources, 
\emph{what} knowledge can be learned and \emph{how} can it be transferred to automatically handle multiple unseen data sources with different value distribution, missing values and new attributes?

To solve this task, human experts typically rely on prior domain knowledge to learn the attribute importance in the seen data sources, and then \emph{transfer} it to match the unseen records based on the similarity of attribute values.
As challenges (\textbf{C1}-\textbf{C3}) lead to different attribute importance, human experts would update their knowledge learned from the seen data sources and adapt to the unseen data sources. 
\dijin{
For example in Figure~\ref{fig:highlight}, when trying to link entities in the red table, the importance of ``Artist'' learned in the seen data sources (blue table) is down-weighted due to the fact that name abbreviation is less informative.
On the other hand, even though it is a rarely-seen attribute in the seen sources, ``Gender'' would be more important because the gender difference between artists naturally leads to non-matching of music records regardless of the fact that entity pairs could share the same ``Title'' (``Hello'') and very similar ``Artist'' values (``A. A.'' and ``A. W.'').
This process, however, is tedious and does not scale to massive unlabeled data involved in real-world entity linkage problems, where large volumes of new data sources continuously arrive. 
}

\dijin{
Following this intuition, we propose \method, a transfer learning framework that leverages both the labeled and massive unlabeled data to train the model for multi-source entity linkage while addressing the aforementioned challenges (\textbf{C1}-\textbf{C3}).
We define the attribute importance in entity linkage as the high-level transferable knowledge and automatically learn it through a proposed attribute-level attention mechanism (\emph{what} to transfer).
In general, as transfer learning aims to transfers knowledge learned from the domain with abundant training data to a related target domain with limited data, the existing works either rely on increasing the labeling volume by introducing the external data (\eg, public knowledge bases)~\cite{zhao2019auto} or reusing the seen training data~\cite{thirumuruganathan2018reuse}.
On the contrary, \method adopts domain adaptation (DA) to jointly update the attention scores for attributes in both the seen and unseen data as the basis for entity linkage (\emph{how} to handle multiple sources), so that the knowledge is adaptive to the continuously incoming data sources.
In addition, the insightful feature importance as transferrable knowledge is explicitly defined by \method to benefit both human interpretation and the performance of learning tasks, which is also different from methods that incorporates the knowledge into pretrained models like ``black-boxes'', such as the contextual word/character embeddings.
While the widely-adopted NLP-based attribute summarization in existing works~\cite{li2020deep,nie2019deep,fu2020hierarchical} could accurately capture the word-level semantics using some pretrained language models or domain knowledge for all attributes, they are too computationally expensive for the practical scenario.
On the contrary, the feature-level attention is much faster to obtain and we claim that the impact of word-level similarity under some attributes is limited and even harmful for model performance if those attributes are not important. 
}

\dijin{
\method follows the real-world scenario and assumes that new data sources come from the same or neighboring domains in batches (\eg, music from different websites). Transferring knowledge between irrelevant domains (\eg, celebrities and products) does not produce meaningful outputs and is out of the scope of this paper.
}
We also propose a series of \method variants for different learning scenarios in practice. 
Our contributions are summarized as follows.
\begin{itemize}
    \item We formulate the problem of MEL in real-world knowledge integration where the incoming data of unseen data sources are associated with missing values, unseen attributes and different value distributions. 
    
        \item We propose a deep transfer-learning framework that learns the attribute-level importance as the high-level knowledge, and incorporates massive unlabeled data across multiple unseen sources through domain adaptation to make it agnostic and transferable.
    \item We apply \method to multi-source entity linkage over both industrial and public datasets,  and show that it achieves at least $5.92\%$ improvement in terms of mean average precision compared to the state-of-the-art deep learning EL methods.
    \end{itemize}


\section{Related Work}
\label{sec:related}

\noindent \textbf{Entity Linkage (EL).} 
Entity linkage has been and continues being a fundamental problem in the field of database, data mining and knowledge integration~\cite{getoor2012entity,kopcke2010frameworks,doan2005semantic}. 
Early works are based on the similarity between entity attributes~\cite{fan2009reasoning,doan2005semantic} through resolving the data conflicts~\cite{dong2009data}, linking relevant attributes through semantic matching or rule reasoning~\cite{singh2017synthesizing}. Techniques such as blocking or hashing are normally applied to merge the candidate entities ~\cite{cohen2002learning}.
The major drawback of these methods is the dependence on prior knowledge as the useful attributes are normally selected through human efforts.
Recently, EL models based on deep neural networks~\cite{mudgal2018deep,joty2018distributed} have been widely studied due to their capability in automatically deriving latent features and promising results in fields such as CV and NLP~\cite{bengio2012deep,luong2015effective,finn2017model}. 
For example, 
DeepER~\cite{joty2018distributed} proposes to leverage RNN to compose the pre-trained word embeddings of tokens within all attribute values, and use them as features to conduct EL as the binary classification task.
DeepMatcher~\cite{mudgal2018deep} also takes the embeddings of attribute words as the input and uses RNN to obtain the attribute similarity representation.CorDel~\cite{wang2020cordel} proposes to compare and contrast the pairwise input records before getting the embeddings so that small but critical differences between attributes can be modeled effectively. 
There are also recent works that formulate entity linkage across different data sources as heterogeneous entity matching~\cite{li2020deep,fu2020hierarchical,nie2019deep}, for example, 
EntityMatcher~\cite{fu2020hierarchical} proposes a hierarchical matching network that jointly match entities in the token, attribute, and entity level.
Ditto~\cite{li2020deep} proposes to leverage the pretrained language model~\cite{joty2018distributed,kenton2019bert,li2020deep} such as BERT or DistilBERT, as well as domain knowledge and data augmentation to improve the matching quality.
The attention mechanisms~\cite{luong2015effective,vaswani2017attention,velivckovic2018graph} are generally adopted by these deep models, where the goal is to improve the linkage performance by highlighting valuable embeddings, \eg, word embeddings within the attributes.
The basis of these above deep models for heterogeneous schema matching is to accurately summarize the attribute words through advanced NLP techniques such as word token-level RNN (with attention) or pretrained language models.
On the contrary, our proposed method does not require sophisticated computation to summarize words in each attribute. \method focuses on the impact of important attributes in matching and explicitly models their importance using the soft attention mechanism as the transferable knowledge.
Such attribute-level importance is agnostic to specific data sources and generalizes better than individual words in the transfer learning paradigm.

\vspace{0.15cm}
\noindent \textbf{Transfer Learning.} 
In the transfer learning scenario, models are trained on a source domain and applied to a related target domain to handle the same or a different task~\cite{pan2009survey,goodfellow2016deep}. 
The specific transferable knowledge that bridges the source and target domain has significant impact to model performance~\cite{ying2018transfer}.
A popular approach is to adapt the pre-trained model for the new task through fine-tuning~\cite{kenton2019bert}, or by adding new functions to specific tasks such as object detection~\cite{he2017mask}.
In terms of EL,
\dijin{TLER~\cite{thirumuruganathan2018reuse} is a non-deep method that reuses and adopts seen data from the source domain to train models for the new domain.
}
Auto-EM~\cite{zhao2019auto} proposes to pre-train models for both attribute-type (\ie, schema) and attribute value matching based on word- and character-level similarity. However, Auto-EM assumes the typed entities are from a single data source and the attributes are seen during training, and thus cannot handle the multi-source scenario with unseen attributes.
A specific type of transductive transfer learning that is most relevant to our work is known as \emph{Domain Adaptation}, where the source and target domain share the same feature space with different distributions~\cite{sun2015survey}, and models are trained on the same task~\cite{wilson2020survey}.
Many well-designed algorithms propose to map the original feature spaces to a shared latent feature space between domains~\cite{duan2012domain,chattopadhyay2012multisource}. 
DeepMatcher+~\cite{kasai2019low} extends DeepMatcher with the combination of transfer learning and active learning to achieve comparable performance with fewer samples. However, this work aims at dataset adaptation rather than the attribute matching, and the focus is not improving the matching performance.
Another line of works proposes to pre-train models on the source and target domain (if labeling available) and then combine them through specific weighting schemes~\cite{schweikert2008empirical}.
The process of applying the trained model to handle previously unseen data is also known as zero-shot learning~\cite{palatucci2009zero}.
Unlike the above approaches, \method explicitly learns feature importance by adapting to the massive unlabeled data from unseen sources as the transferable knowledge for the multi-source EL task.

\section{Preliminaries}
\label{sec:prelim}

In this section, we first formally define the problem, and then provide several key notions relevant to our proposed solution. Symbols and notations used in this paper are listed in Table~\ref{table:symbols}.

\begin{table}[t!]
\centering
\caption{Summary of notation}
\vspace{-.3cm}
\setlength{\tabcolsep}{5pt} \resizebox{.96\columnwidth}{!}{
\begin{tabular}{@{}l p{7.5cm}}
\toprule
  \textbf{Symbol} & \textbf{Definition} 
\\ 
\midrule
$\mathcal{A} = \{A_j\}$ & a set of pre-defined textual attributes (data source schema)  \\
$r, r[A]$ & an entity record and the value (word tokens) of attribute $A$ \\
$\mathcal{D}_S, \mathcal{D}_T$ & source and target domain, respectively \\
$(r, r^{\prime})_{S/T}$ & an entity pair in the source and target domain, respectively \\
$S, S^{\prime}$ & set of data sources in general \\
$r^{\ast}$ & the data source that record $r$ is sampled from \\
$\mathcal{D}^{\ast}$ & set of data sources in a domain, \eg, $\mathcal{D}_S^{\ast}=\{r^{\ast}\}_{r\in\mathcal{D}_S}$ \\
$F$ & the number of relational features, $F=2|\mathcal{A}|$\\
$\mathbf{x}, y$ & $H$-dim latent feature vector of an entity pair and its label \\
$\mathbf{h}_j$ & $D$-dim token embedding of feature $j$  \\
$f$ & attention embedding function $\mathbb{R}^{D\times F} \rightarrow \mathbb{R}^{F}$ \\
  \bottomrule
\end{tabular}
}
\label{table:symbols}
\end{table}

\subsection{Problem Definition}
\label{sec:problem_statement}

An entity record is collected from a specific data source such as a website or a database, and is identified by its attributes. For example, a song record $r=$ (``Sweet Caroline'', ``Neil Diamond'', ``USA'') is specified by the attributes $\mathcal{A}=\{\texttt{title, artist, country}\}$. We start with the formal definition of entity linkage.

\begin{problem}[EL: Entity Linkage]
Given two entity records $r$ and $r^{\prime}$ associated with the same set of attributes $\mathcal{A}$ (schema), entity linkage aims to predict if $r$ and $r^{\prime}$ refers to the same real-world entity.

\label{def:problem-statement-pre}
\end{problem}

In this paper, we conduct analysis based on entity pairs $(r, r^{\prime})$ instead of individual entity records. 
We now define the MEL problem, which is related to  heterogeneous entity matching\footnote{
In MEL, the entities come from different data sources, and thus there may be new or missing attributes. On the other hand, in heterogeneous entity matching, the schemas are heterogeneous (i.e., they have different attributes, which may not be aligned) and the entities do not necessarily come from different data sources.}~\cite{nie2019deep,fu2020hierarchical}.

\begin{problem}[MEL: Multi-source Entity Linkage]
Given the labeled entity pairs $\{(r, r^{\prime})\}_{\text{seen}}$ from a limited set of data sources $\mathcal{S}$ 
where each entity record $r$ is associated with attributes $\mathcal{A}$, and previously unseen pairs $\{(r, r^{\prime})\}_{\text{unseen}}$ from the new data sources $\mathcal{S}^{\prime}$ with attributes $\mathcal{A}^{\prime}$,
multi-source entity linkage aims to predict if 
each pair in $\{(r, r^{\prime})\}_{\text{unseen}}$ represents the same real-world entity, where $(r, r^{\prime})_{\text{unseen}}\in(\mathcal{S}\bigtimes\mathcal{S}^{\prime}) \cup (\mathcal{S}^{\prime}\bigtimes\mathcal{S}^{\prime}) $, $|\mathcal{S}^{\prime}| > |\mathcal{S}|$.
Since $\mathcal{S}\neq\mathcal{S}^{\prime}$, certain attributes in $\mathcal{A}^{\prime}$ could be missing (\textbf{C1}), new (\textbf{C2}), or associated with values from different distributions (\textbf{C3}), and thus $\mathcal{A}\neq\mathcal{A}^{\prime}$.

\label{def:problem-statement}
\end{problem}

The key notion in Problem~\ref{def:problem-statement} that is different from Problem~\ref{def:problem-statement-pre} is that the linkage task is conducted on entity pairs sampled from a wider range of data sources than the labeled data used to train the model (ten or hundred orders of magnitude more in reality).
Back to the example shown in Figure~\ref{fig:highlight}, while the trained model could make perfect prediction based on ``Artist'' only, it would fail to handle new records because the attribute ``Artist'' has missing or abbreviated values that contain less info.
Moreover, the new data sources contain a rarely seen or unseen attribute (``Gender'').
This issue can be addressed by aligning the union of ontology  $\mathcal{A}\cup\mathcal{A}^{\prime}$ with blank ``dummy'' attributes.
Based on our definition, a solution to MEL should be able to (\textbf{G1}) make use of the massive unlabeled data from the new sources, and (\textbf{G2}) further improve the linkage performance by leveraging a few labeled record pairs from the new sources, if available (\ie, an additional support set).

\subsection{Terminology}
Here we discuss the necessary terminology of our framework,.

\begin{definition}[Source \& target domain]
The source domain $\mathcal{D}_S$ refers to a set of labeled entity pairs $\{(r, r^{\prime})\}$ sampled from limited data sources that the model is trained on.
The target domain $\mathcal{D}_T$ refers to the set of unlabeled pairs where each pair has at least one entity sampled from the data sources unseen in $\mathcal{D}_S$.

\label{def:source_target_domain}
\end{definition}

For clarity, we use the superscript $^{\ast}$ to indicate the data source(s) of a record/domain.
Following Definition~\ref{def:source_target_domain}, the seen and unseen set of data sources in Problem~\ref{def:problem-statement} are formulated as $\mathcal{S} = \mathcal{D}^{\ast}_S$ and $\mathcal{S}^{\prime} = \mathcal{D}^{\ast}_T$.
Besides, given a pair in the target domain, it could either contain one entity sampled from the seen data sources in $\mathcal{D}^{\ast}_S$ and the other one from the unseen, \ie, $(r, r^{\prime})_T\in\mathcal{D}^{\ast}_S\bigtimes \mathcal{D}^{\ast}_T$, or it has both entities sampled from the completely unseen data sources, \ie, $(r, r^{\prime})_T\in\mathcal{D}^{\ast}_T\bigtimes \mathcal{D}^{\ast}_T$.
In both cases, achieving \textbf{G1} requires data in $\mathcal{D}_T$.
To achieve \textbf{G2}, we introduce the support set.

\begin{definition}[Support set]
The support set $\mathcal{S}_U$ refers to a small set of labeled entity pairs sampled from the same set of data sources as the target domain $\mathcal{D}^{\ast}_T$. It has at least one data source that is not contained in $\mathcal{D}^{\ast}_S$.
\label{def:support_set}
\end{definition}
The support set corresponds to the real-world scenario that a few newly incoming entity pairs are well-labeled (\eg, on-the-fly human annotation).
Thus, entity pairs in $\mathcal{D}_S$, $\mathcal{D}_T$, as well as $\mathcal{S}_U$ are all required to achieve \textbf{G2} for MEL.

\section{Proposed framework}
\label{sec:method}

\begin{figure}[t!]
\centering
\includegraphics[width=0.96\linewidth]{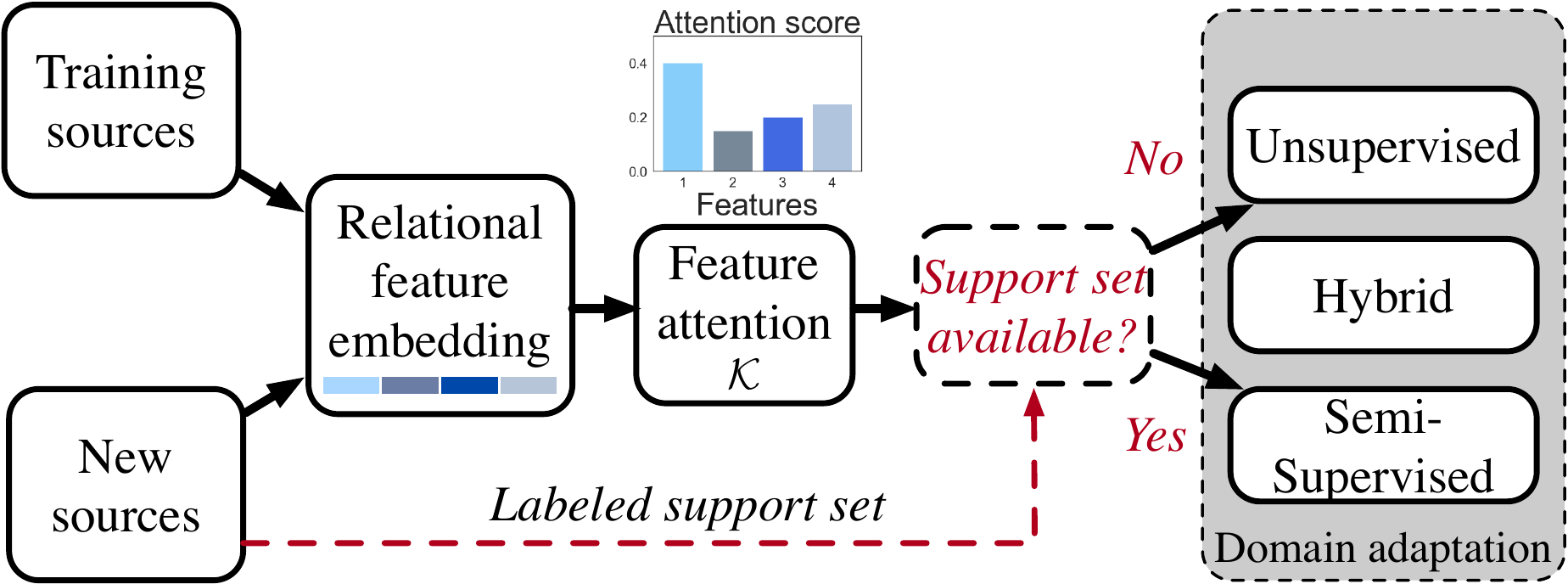}
\caption{
Overview. 
\method first embeds attributes for records from both the source and target domain to derive the feature representations, and uses the feature attention function to get the attention scores (importance) as the transferable knowledge $\mathcal{K}$. Then, depending on the availability of the labeled support set, \method uses $\mathcal{K}$ and performs either the unsupervised or semi-supervised manner of domain adaptation for MEL.
}
\label{fig:overview}
\vspace{-.3cm}
\end{figure}

\dijin{
We propose \method to address Problem~\ref{def:problem-statement}, a deep framework that learns attribute importance as the transferable knowledge $\mathcal{K}$ (Section~\ref{sec:method-formulation}), and adapt it to multiple data-sources via domain adaptation .
\method first extracts the contrastive relational features of entity pairs to derive the embeddings (Section~\ref{sec:method-preprocess}). Then, by using the proposed attention embedding function $f$, \method projects features from $\mathcal{D}_S$ and $\mathcal{D}_T$ into the same attention space (Section~\ref{sec:method-feature-attention}), and jointly learns the feature importance for data sources in both $\mathcal{D}_S^{\ast}$ and $\mathcal{D}_T^{\ast}$. This process is conducted in an unsupervised or supervised domain adaptation manner (Section\ref{sec:method-variants}), depending on the real-world scenario.
The overview is depicted in Figure~\ref{fig:overview}.
}

\subsection{Formulation}
\label{sec:method-formulation}

In transfer learning, the generic transferable knowledge $\mathcal{K}$ is key to adapt the model trained on the source domain to the target domain.
We denote our domain adaptation solution to MEL as the following binary classification task.
\begin{equation}
    y=M(\mathcal{K}, (r, r^{\prime})) \in \{0, 1\}
    \label{eq:problem-statement}
\end{equation}
where $M$ represents the deep model that generates the binary prediction $y$ for the entity pair $(r, r^{\prime})\in\mathcal{D}_T$, where $1$ and $0$ indicate the matching and non-matching, respectively.
As mentioned in Section~\ref{sec:problem_statement}, the key difference between $\mathcal{D}_S$ and $\mathcal{D}_T$ lies in the difference in data sources, therefore $\mathcal{K}$ should be data-source agnostic to address \textbf{(C1)}-\textbf{(C3)}.
To ensure $\mathcal{D}_T$ shares the same feature space as $\mathcal{D}_S$ (the prerequisite for domain adaptation), \method first aligns the ontology so that data sources $\mathcal{D}^{\ast}_S$ and $\mathcal{D}^{\ast}_T$ share the same attribute schema, but the attribute values (word tokens) can vary significantly. 
By doing so, entity records reveal the following properties that correspond to the aforementioned challenges: (\textbf{C1})
entity records in the source/target domain contain missing values, \ie, $r[A]=$``'' (empty string) for $r\in\mathcal{D}_S \cup \mathcal{D}_T$, (\textbf{C2}) certain attribute values are completely missing for records in $\mathcal{D}_S$, \ie, $r[A_j]=$``'' for $r\in\mathcal{D}_S$, but not in $\mathcal{D}_T$, and (\textbf{C3})
rich texts under some attributes in $\mathcal{D}_S$ but sparse in $\mathcal{D}_{T}$ or vice versa.

\subsection{Feature Representation}
\label{sec:method-preprocess}

Given entity pairs $(r, r^{\prime})$ with the aligned attributes $\mathcal{A}$, \method leverages the attention mechanism to learn the importance of each textual attribute $A\in \mathcal{A}$ as the generic knowledge for transfer learning. 
However, instead of computing the attribute importance directly, \method parses each attribute $A$ into $2$ contrastive relational features, which are word tokens shared by $r$ and $r^{\prime}$, and word tokens that only appear in one record but not the other. This is because the similarity or uniqueness of attribute between $r$ and $r^{\prime}$ gives independent and complementary evidence for linkage \cite{wang2020cordel}.
Taking the attribute $A=$``music version'' as an example, a pair of music recordings sharing the same word (\ie, ``original'' or ``remix'') is not as strong an identifier for matching as it would be for non-matching if one recording is ``original'' while the other is ``remix''.
In addition, looking into both the similarity and uniqueness in attribute $A$ between entities would enrich the feature space and facilitate training the deep model.
We describe the 2 contrastive relational features of an attribute $A$ as follows.
\begin{equation}
    \begin{cases}
        sim(A) &= \{w\} \text{ for } w\in \{r[A] \cap r^{\prime}[A]\} \\
        uni(A) &= \{w\} \text{ for } w\in \{r[A] \cup r^{\prime}[A] - r[A] \cap r^{\prime}[A]\}
    \end{cases}
\end{equation}
where $w$ is the word token in attribute $r[A]$.
For clarity, we uniformly denote shared/unique tokens $sim(A)/uni(A)$ as ``features'' that contribute independently to entity linkage. Clearly, there are $F=2|\mathcal{A}|$ features for a pair of entities.
To summarize the feature representation, \method  sums up the embeddings of the cropped word tokens~\cite{joulin2017bag,vaswani2017attention,mudgal2018deep} without using more sophisticated operations.
The embeddings can be obtained using any pretraining language model, such as    BERT~\cite{kenton2019bert} or Fasttext~\cite{joulin2017bag}.
For clarity, we use $i$ as the index of entity pairs and $j$ as the index of features.
Thus, the token embedding vector of an entity pair $(r, r^{\prime})$ is denoted as:
\begin{align}
    \mathbf{h} &= [\mathbf{h}_{1}, \mathbf{h}_{2}, \cdots, \mathbf{h}_{F}] \nonumber\\
    &= [\text{emb}(sim(A_j)), \text{emb}(uni(A_j))] \text{ for } j=1, \cdots, |\mathcal{A}|
\end{align}

\begin{figure}[t!]
\vspace{-.1cm}
\centering
\includegraphics[width=0.92\linewidth]{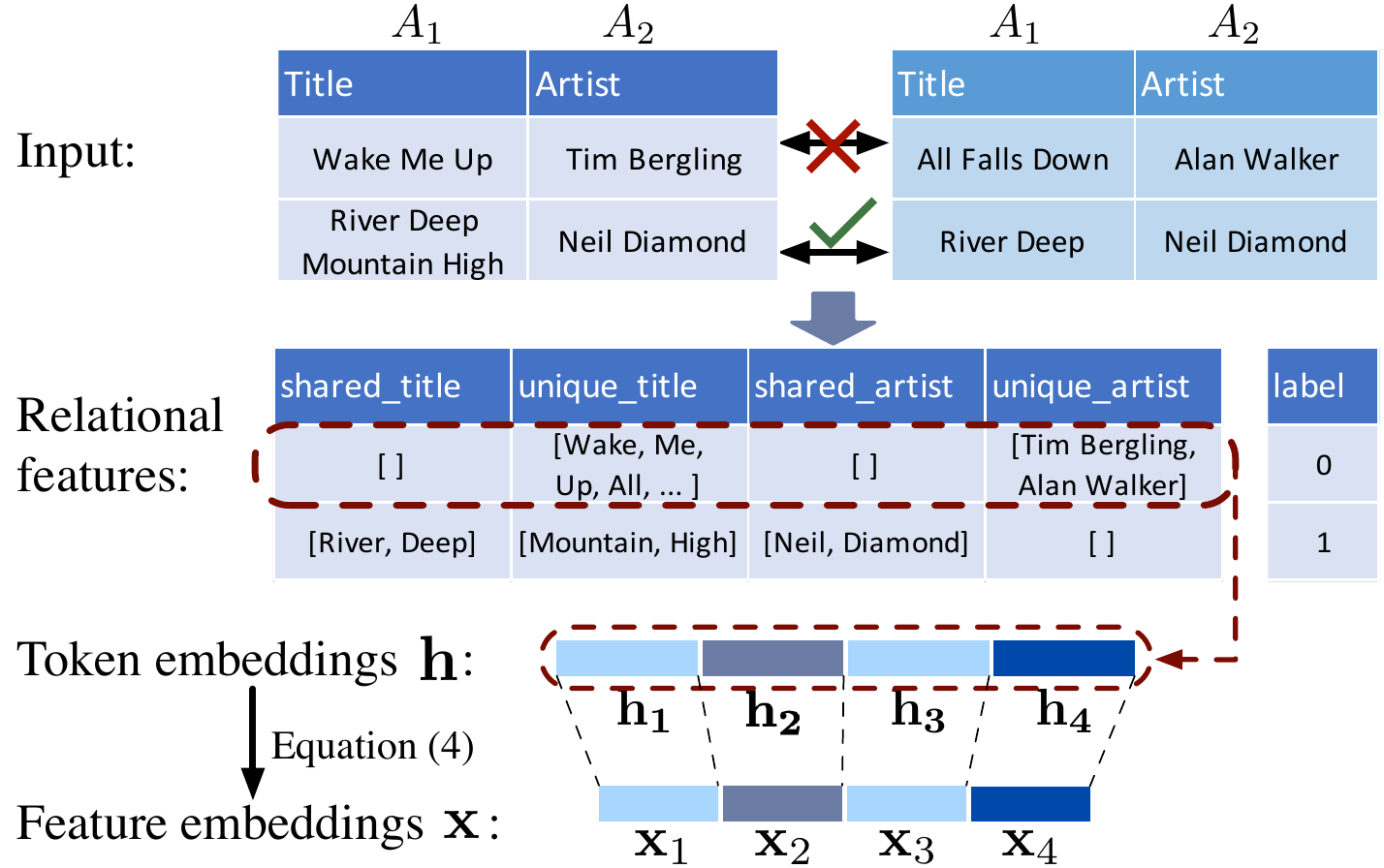}
\vspace{-0.2cm}
\caption{
    \method processes $1$ attribute $A$ as $2$ relational features (\ie, $sim(A)$ and $uni(A)$). In this example, $F=4$ features are generated from $|\mathcal{A}|=2$ attributes (\ie, ``Title'' and ``Artist'').
        The empty word tokens are embedded as the fixed normalized non-zero vector to form $\mathbf{h}$ (red dashed box).
    The feature embedding $\mathbf{x}$ is obtained through non-linear affine transformation of the token embedding $\mathbf{h}$ (Equation~\eqref{eq:affine}).
    Each feature assumes to contribute independently to predict the linkage. 
}
\label{fig:preprocess}
\vspace{-.4cm}
\end{figure}

By doing so, we denote the entity pairs $(r, r^{\prime})$ as $F$ textual embedding features ($F=2|\mathcal{A}|$) for matching. The complete process is depicted in Figure~\ref{fig:preprocess}.
Besides, \method leverages per-feature non-linear affine transformation to project the word embeddings to get the latent feature $\mathbf{x}$:
\begin{align}
    \mathbf{x} = [\mathbf{x}_1, \mathbf{x}_2, \cdots, \mathbf{x}_F] =  [\sigma(\mathbf{V}_j\mathbf{h}_j + \mathbf{b}_j)] \text{ for } j=1, \cdots, F
    \label{eq:affine}
\end{align}
where $\mathbf{V}_j^{H\times D}$ is the learnable weight matrix, $\mathbf{b}_j^{H}$ is the learnable bias vector, and $\sigma$ denotes the non-linear activation function (\eg, Relu).
With this representation, Equation~\eqref{eq:problem-statement} can be rewritten as: 
$y=M(\mathcal{K}, \mathbf{x})\in\{0, 1\}$.
Next we discuss how \method learns feature importance \footnote{In this paper, we compute the feature attention as the transferable knowledge, feature importance.} as the transferable knowledge $\mathcal{K}$.

\subsection{Feature Attention Embedding}
\label{sec:method-feature-attention}

Given a pair of entities denoted through $F$ features, \method defines the energy score of feature $j$ as $e_{j} = a(\mathbf{Wx}_{j})$,
where $\mathbf{x}_{j}$ is the $H$-dimensional representation of latent feature $j$, $\mathbf{W}^{H^{\prime} \times H}$ is a shared linear transformation, and
$a$ represents the attention mechanism $\mathbb{R}^{H^{\prime}} \rightarrow \mathbb{R} $, as a single-layer neural network (parameterized with $\mathbf{a}$).
\method allows each feature to attend to the label $y$ independently and computes coefficients using the softmax function such that the normalized scores are comparable across all features. Formula in Equation~\eqref{eq:self_attention} computes the attention score of feature $j$:
\begin{align}
    g(\mathbf{x}_{j}) &= \text{softmax}_j(e_{j}) =\frac{\exp{(\mathbf{a}^T \tanh{(\mathbf{Wx}_{j})}})}
    {\sum_{k=1}^{F}\exp{(\mathbf{a}^T \tanh{(\mathbf{Wx}_{k})}})}
    \label{eq:self_attention}
\end{align}

Note that Equation~\eqref{eq:self_attention} only generates the scalar attention score of feature $j$ for an input vector $\mathbf{x}$. To compute the scores of all features, we introduce the attention embedding function $f$ that learns attention scores of all $F$ features as follows.
\begin{align}
    f(\mathbf{x}) = f([\mathbf{x}_{1}, \mathbf{x}_{2}, \cdots, \mathbf{x}_{F}]) = [g(\mathbf{x}_1), g(\mathbf{x}_2), \cdots, g(\mathbf{x}_F)]
    \label{eq:fx}
\end{align}

In Equation~\eqref{eq:fx}, all features share the same $\mathbf{W}$ and $\mathbf{a}$ to compute the attention scores. We denote $f(\mathbf{x})_j=g(\mathbf{x}_j)$, and $\sum_{j=1}^{F}f(\mathbf{x})_j = 1$.
\method takes the generated feature importance vector $f(\mathbf{x})$ as the transferable knowledge $\mathcal{K}$ for the entity pair $(r, r^{\prime})$, \ie, $\mathcal{K} = f(\mathbf{x})$.

In the learning process, \method feeds the feature representation coupled with its attention score to a 2-layer feed-forward neural network $\Theta$ to perform the binary classification task:
\begin{equation}
    \hat{y} = \Theta(\sigma(f(\mathbf{x}) \odot \mathbf{x})) = \Theta([\sigma(g(\mathbf{x}_1)\cdot \mathbf{x}_{1}),  \cdots, \sigma(g(\mathbf{x}_F)\cdot \mathbf{x}_{F})])
    \label{eq:nn}
\end{equation}
where $\odot$ denotes the element-wise multiplication, $\sigma$ denotes the non-linear activation (\eg, Relu) and $\hat{y}$ denotes the inference score for matching. 
\method uses the same attention mechanism to handle all records in the training and leverages the cross-entropy loss to update the shared parameters $\mathbf{W}$, $\mathbf{a}$, as well as the learnable $\mathbf{V}$, $\mathbf{b}$ through back-propagation:
\begin{equation}
    L_{base} = -\frac{1}{N}\sum^{N}_{i=1}y_i\log{\hat{y}_i} + (1-y_i)\log(1-\hat{y}_i)
    \label{eq:loss_1}
\end{equation}
where $y_i$ denotes the label $\{0, 1\}$. To ensure that all learnable parameters can be updated correctly, \method initializes the missing attribute values (incurred by challenge \textbf{C1, C2}) with a fixed normalized non-zero vector.

\begin{figure}[t!]
\vspace{-.1cm}
\centering
\includegraphics[width=0.8\linewidth]{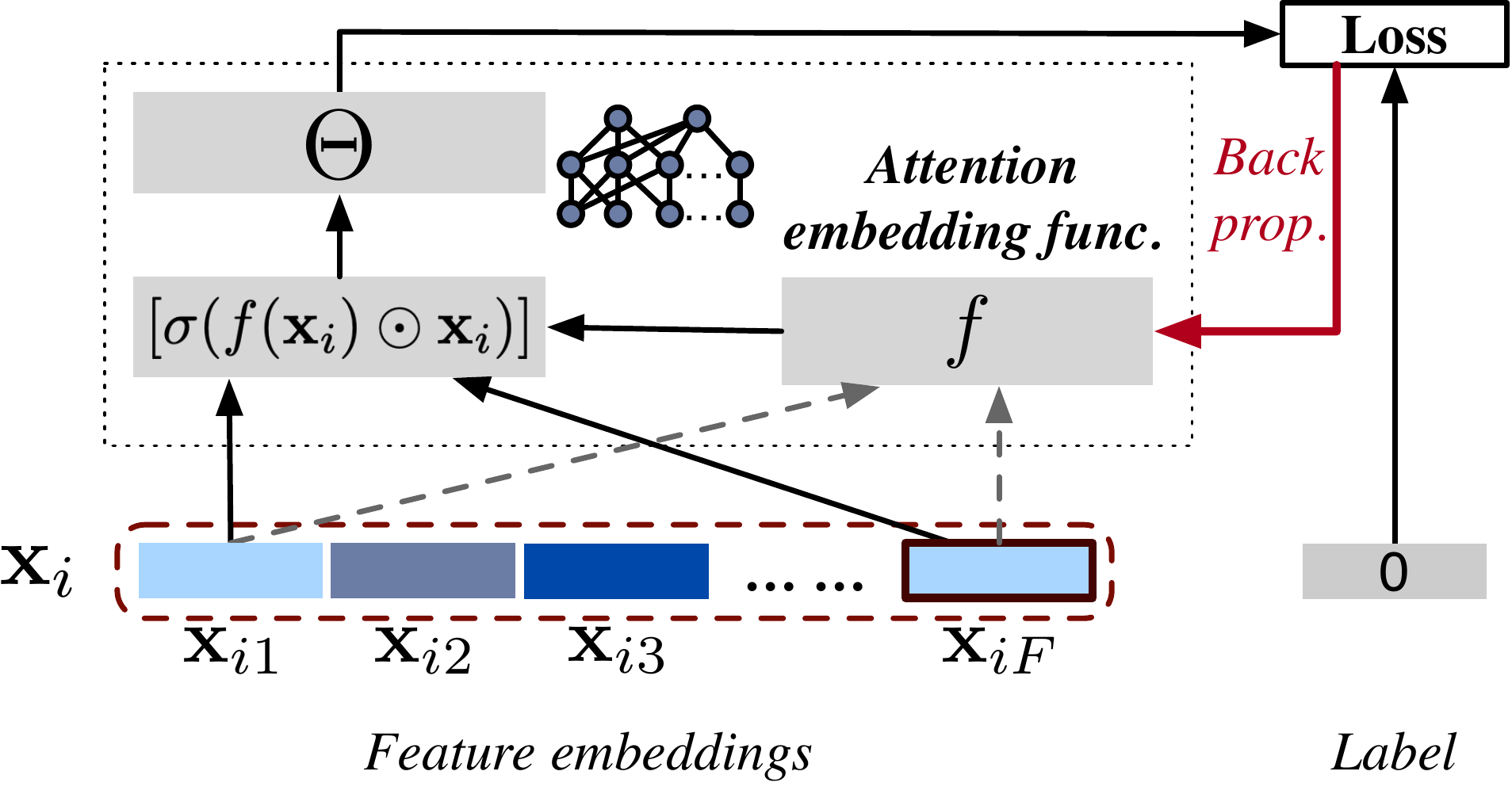}
\vspace{-0.3cm}
\caption{
\method-base architecture that updates $f$ via labeled data in $\mathcal{D}_S$. \method-base first computes the attention vector $f(\mathbf{x}_i)$ for the $i-$th entity pair (dashed line), and then compose it with the feature embeddings (solid line) as the input to the neural network $\mathbf{\Theta}$.
}
\label{fig:architecture_v1}
\vspace{-.3cm}
\end{figure}

We name this solution \textbf{\method-base} as it learns $f$ through the labeled data in $\mathcal{D}_S$ and illustrate the architecture in Figure~\ref{fig:architecture_v1}.
The attribute importance learned under the supervision of labeled data in $\mathcal{D}_S$ will be carried over to the unseen data sources and may not generalize well as there is always new data from seen or unseen sources with different distributions (\textbf{C3}) in MEL. 
Next we discuss how \method adopts $\mathcal{D}_T$ sampled from multiple data sources to alleviate this issue and make $\mathcal{K}$ data-source agnostic.

\subsection{Domain Adaptation-based Variants}
\label{sec:method-variants}

Based on \method-base, we propose three variants that leverage domain adaptation to handle different learning scenarios. 

\subsubsection{Unsupervised Domain Adaptation}  \label{sec:method-zsl}

Our first idea is to adjust the learned attribute importance according to new distribution of unlabeled data.
In Equation~\eqref{eq:fx}, the attention embedding function $f$ contains the shared attention mechanism $a$ parameterized by weight vector $\mathbf{a}$ and the shared transformation matrix $\mathbf{W}$. It only takes the feature embeddings $\mathbf{x}$ as input to compute the attention scores.
Since $\mathbf{W}$ and $\mathbf{a}$ are shared across the input data, the attention score vector $f(\mathbf{x})$ can be seen as projecting the input feature embeddings $\mathbf{x}$ into a hyper-plane that is parameterized by $\mathbf{W}$ and $\mathbf{a}$.
Without introducing extra information such as entity pair labeling, we can project data from $\mathcal{D}_T$ into the same space as $\mathcal{D}_S$, and it holds as long as the ontology of the unlabeled data aligns with the labeled data, \ie, identical attribute schema between $\mathcal{D}_S$ and $\mathcal{D}_T$.

Therefore, \method uses the KL divergence to measure the attention score distribution difference between the source and target domain as the regularization term to train the model.
The loss is defined in Equation~\eqref{eq:loss_un}.
At a specific iteration in the training, \method uses the up-to-date $f$ to project data from both $\mathcal{D}_S$ and $\mathcal{D}_T$ into the same feature attention space. 
Then, \method updates $\mathbf{W}$ and $\mathbf{a}$ so that not only the cross-entropy loss introduced in Equation~\eqref{eq:loss_1} is minimized, but also the KL divergence between feature attention distributions for  $\mathcal{D}_S$ and $\mathcal{D}_T$.
In this way, feature importance for entity records in $\mathcal{D}_S$ is jointly updated with records sampled from a wider range of data sources in $\mathcal{D}_T$, and thus being agnostic to previously unseen data sources that have significantly different value distribution (\textbf{C3}).
\begin{equation}
        L_{\text{un}} = (1-\lambda) L_{\text{base}} + \lambda L_{\text{target}}
    \label{eq:loss_un}
\end{equation}
where $\lambda$ is the hyperparameter that balances between $L_{base}$ and $L_{\text{target}}$. $\lambda$ also measures the amount of adaptation to the target domain $\mathcal{D}_T$.
$L_{\text{target}}$ is given as follows.
\begin{equation}
    L_{\text{target}} = 
    \text{KL}(f(\mathbf{x}), \bar{f}(\mathbf{x}^{\prime})) = \sum^{|\mathcal{D}_S|}_{i=1}\sum_{j=1}^{F}\bar{f}(\mathbf{x}^{\prime})_{j}\log(\frac{\bar{f}(\mathbf{x}^{\prime})_{j}}{f(\mathbf{x}_{i})_j})
    \label{eq:loss_un_target}
\end{equation}
where $\bar{f}(\mathbf{x}^{\prime})_j=\frac{1}{|\mathcal{D}_T|}\sum_{\mathbf{x}^{\prime}_i\in\mathcal{D}_T}f(\mathbf{x}^{\prime}_i)_j$, which represents the attention score for feature $j$ averaged over the unlabeled data. 
$\mathbf{x}$ and $\mathbf{x}^{\prime}$ denote the feature vector in the source and target domain, respectively, and $f(\mathbf{x}_i)_{j}$ denotes the importance of the $j$-th feature in the $i$-th entity pair.
In practice, \method adopts batch learning to improve the training efficiency, \ie, minimizing the loss per batch instead of iterating through all records in the data. 
The unlabeled data could also come in batches, which makes $\bar{f}(\mathbf{x}^{\prime})$ be the attention vector averaged over the batched unlabeled data instead of all in the target domain. By default, the batches are sampled randomly.

\begin{figure}[t!]
\centering
\includegraphics[width=0.92\linewidth]{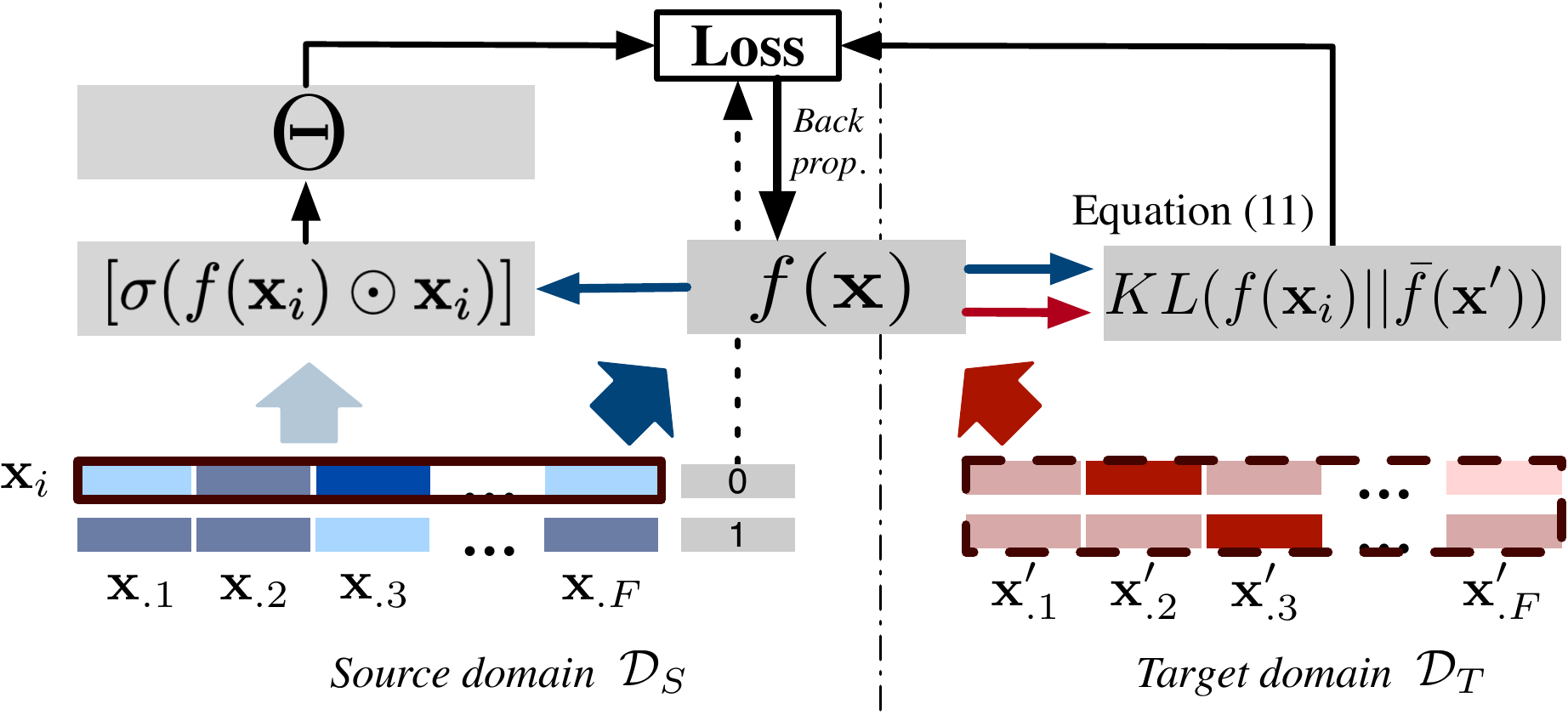}
\vspace{-0.3cm}
\caption{
\methodA architecture that attempts to align the $i$-th entity pair $f(\mathbf{x}_i)$ in $\mathcal{D}_S$ (solid box) with the averaged $f(\mathbf{x^{\prime}})$ (dashed box) in $\mathcal{D}_T$. $\mathbf{x}_{.j}$ and $\mathbf{x}^{\prime}_{.j}$ ($j=1, \cdots, F$) denote the $j$-th feature in general from $\mathcal{D}_S$ and $\mathcal{D}_T$, respectively.
}
\label{fig:architecture}
\vspace{-.3cm}
\end{figure}

\dijin{
We name this solution \textbf{\methodA} as it is based on unsupervised domain adaption without using any labeled data in $\mathcal{D}_T$ and performs linkage in the zero-shot manner. This model also follows the design pattern in~\cite{ganin2015unsupervised}. Figure~\ref{fig:architecture} depicts the architecture and the algorithm is given in Algorithm~\ref{algo:method-un}.
}
Line~\ref{alg1-line-affine-start}-\ref{alg1-line-affine-end} project the affine transformation of entity pairs from both $\mathcal{D}_S$ and $\mathcal{D}_T$. Line~\ref{alg1-line-target} computes $\bar{f}(\mathbf{x}^{\prime})$, the attention vector averaged over entity pairs in the target domain. Line~\ref{alg1-line-attention-start}-\ref{alg1-line-attention-end} computes each attention vector in the sampled batch $f(\mathbf{x}_i)$ and adapt it to $\bar{f}(\mathbf{x}^{\prime})$ to compute the loss $L_{\text{target}}$. \method minimizes both the inference loss $L_{\text{base}}$ and $L_{\text{target}}$ to train the parameters in $f$ and form the transferable knowledge $\mathcal{K}=f(\mathbf{x}_i)$ for $\mathbf{x}_i\in\mathcal{D}_T$ (Line~\ref{alg1-line-knowledge}). Line~\ref{alg1-line-reference-start}-~\ref{alg1-line-reference-end} denote the inference.

\setlength{\floatsep}{20pt}
\setlength{\textfloatsep}{8pt}
\begin{algorithm}[t!]
\caption{\methodA}
\begin{algorithmic}[1]
\Ensure $\mathcal{D}_{S}=\{(\mathbf{h}_i, y_i)\}$, $\mathcal{D}_{T}=\{\mathbf{h}_i\}$, $\lambda$, batch size $B$

\Require Predicted $\hat{y_i}$ for $\mathbf{h}_i \in\mathcal{D}_{T}$, updated $\mathbf{a, W}$


\State Initialize $\mathbf{a, W}$ and $\mathbf{V, b}$

\Loop { training epochs}
    \For{$\mathbf{h}\in\mathcal{D}_S\cup\mathcal{D}_T$} \label{alg1-line-affine-start}
        \State Form $\mathbf{x}$ with $\mathbf{V, b}$\Comment{Eq.~\eqref{eq:affine}}
    \EndFor \label{alg1-line-affine-end}
    
    \State $\bar{f}(\mathbf{x}^{\prime}) \leftarrow \frac{1}{|\mathcal{D}_{T}|}\sum_{\mathbf{x_i}\in\mathcal{D}_{T}}f(\mathbf{x_i})$  \label{alg1-line-target}
    \State $J \leftarrow 0$ \Comment{Initialize loss}
    \State $\mathcal{S}_{\text{batch}} \leftarrow \text{RANDOMSAMPLE}(\mathcal{D}_{S}, B)$


    \For{$(\mathbf{x},y) \in \mathcal{S}_{\text{batch}}$}  \label{alg1-line-attention-start}
        \State  $L_{\text{un}} \leftarrow (1-\lambda)L_{\text{base}} + \lambda L_{\text{target}}$ \Comment{Eq.~\eqref{eq:loss_un}}
        \State $J \leftarrow J + \nabla L_{\text{un}}$ \Comment{Update $\mathbf{a}, \mathbf{W}, \mathbf{V}, \mathbf{b}$}
        
    \EndFor \label{alg1-line-attention-end}
\EndLoop

\State Form $\mathbf{x}, f$ with updated $\mathbf{a, W}$, $\mathbf{V, b}$ \Comment{Eq.~\eqref{eq:fx}} \label{alg1-line-knowledge}

    
\State $\hat{\mathbf{y}} \leftarrow \emptyset$ \label{alg1-line-reference-start-before}
\For{$\mathbf{x}_i\in \mathcal{D}_{T}$} \label{alg1-line-reference-start}
    \State $\hat{\mathbf{y}}_i \leftarrow \Theta(\sigma(f(\mathbf{x}_i)\odot \mathbf{x}_i))$
\EndFor \label{alg1-line-reference-end}

\State \textbf{return} $\hat{\mathbf{y}}$, $\mathbf{a, W}$

\end{algorithmic}
\label{algo:method-un}
\end{algorithm}

\vspace{-0.2cm}
\subsubsection{Semi-supervised Domain Adaptation}
\label{sec:method-fsl}

In practice, a small number of labels may be available for the entity pairs coming from the target domain (\eg, through on-the-fly human annotation). Entity pairs in this support set $\mathcal{S}_{U}$ are sampled from the wide range of data sources and provide clues about the data characteristics of the target domain.
To leverage this set of labeled data (\textbf{G2}), \method updates the attention embedding function $f$ under the supervision of $\mathcal{S}_{U}$ so that the projected feature attention vectors of entity pairs in $\mathcal{D}_S$ could match to those in $\mathcal{S}_U$. 
For this purpose, \method computes the centroid of the positive entity pairs in $\mathcal{D}_S$ as follows:
\begin{equation}
    \mathbf{c}_{\mathcal{D}_S}^{+} = \frac{1}{|\mathcal{D}_{S}|}\sum_{(\mathbf{x}^{+}_i, y^{+}_i)\in\mathcal{D}_{S}} f(\mathbf{x}_i)
    \label{eq:centroids}
\end{equation}
The centroid of the negative pairs can be computed in a similar way
with negative samples.
Intuitively, entity pairs from the data sources unseen in $\mathcal{D}^{\ast}_S$ are more important in adaptation than those from the seen sources, and should be highlighted.
\method measures such difference through the Euclidean-distance between $f(\mathbf{x})$ and $\mathbf{c}_{\mathcal{D}_S}$, as the deviating attention vectors are more likely to correspond to unseen data sources in the projected space.
In the loss function shown in Equation~\eqref{eq:loss_support}, we compare the distance $d(f(\mathbf{x}), \mathbf{c}_{\mathcal{D}_S})$ with the ``mean distance to cluster centroids'' to give higher weights to entity pairs in $\mathcal{S}_U$ that are deviating from seen data sources.
\begin{equation}
    L_{\text{support}} = \sum_{y_i=1}\frac{d(f(\mathbf{x}^{+}_i), \mathbf{c}_{\mathcal{D}_S}^{+})}{\bar{d}^{+}_{\mathcal{D}_S}}\log \hat{y_i} + \sum_{y_i=0}\frac{d(f(\mathbf{x}^{-}_i), \mathbf{c}_{\mathcal{D}_S}^{-})}{\bar{d}^{-}_{\mathcal{D}_S}}\log (1-\hat{y_i})
    \label{eq:loss_support}
\end{equation}
where $d$ denotes the Euclidean distance, $\bar{d}^{+/-}$ represents the mean distance for all positive/negative pairs in $\mathcal{D}_{S}$ to the corresponding centroid.
Thus, by integrating $L_{support}$ with $L_{base}$, the updated loss of \method in the supervised setting is denoted as follows:
\begin{equation}
    L_{\text{ssl}} = L_{\text{base}} + \phi L_{\text{support}}
    \label{eq:loss_2}
\end{equation}
where $\phi \in (0, 1]$ is a hyperparameter that controls the impact of the labeled support set.
The training process updates not only parameters in the neural network $\Theta$ for better classification performance, but also the attention embedding function $f$ so that the projected positive and negative feature attentions are matched closer. In this process, feature importance from the new data sources unseen in $\mathcal{D}_S^{\ast}$ can be incorporated to update the centroids $\mathbf{c}^{+/-}$ in the supervised manner.
We name this solution \textbf{\methodB} as it uses a few labeled data in $\mathcal{D}_T$, and depicts the process in Algorithm~\ref{algo:method-ssl}. 
Particularly, line~\ref{alg2-line-attention-start}-~\ref{alg2-line-attention-end} denote the training process of $f$ to minimize the loss $L_{\text{base}}$, and line~\ref{alg2-line-attention-support-start}-~\ref{alg2-line-attention-support-end} denote the process of further training under the supervision of labeled samples in $\mathcal{S}_U$.

\setlength{\textfloatsep}{8pt}
\begin{algorithm}[t!]
\caption{\methodB}
\begin{algorithmic}[1]
\Ensure $\mathcal{D}_{S}=\{(\mathbf{h}_i, y_i)\}$, $\mathcal{S}_{U}=\{(\mathbf{h}_i, y_i)\}$, $\mathcal{D}_{T}=\{\mathbf{h}_i\}$, $\phi$, $B$

\Require Predicted $\hat{y_i}$ for $\mathbf{h}_i \in\mathcal{D}_{T}$, updated $\mathbf{a, W}$


\State Initialize $\mathbf{a, W}$ and $\mathbf{V, b}$

\Loop { training epochs}
    \For{$\mathbf{h}\in\mathcal{D}_S\cup\mathcal{D}_T$} \label{alg2-line-affine-start}
        \State Form $\mathbf{x}$ with $\mathbf{V, b}$\Comment{Eq.~\eqref{eq:affine}}
    \EndFor \label{alg2-line-affine-end}

    \State $J \leftarrow 0$ \Comment{Initialize loss}
    \State $\mathcal{S}_{\text{batch}} \leftarrow \text{RANDOMSAMPLE}(\mathcal{D}_{S}, B)$


    \For{$(\mathbf{x},y) \in \mathcal{S}_{\text{batch}}$} \label{alg2-line-attention-start}
        
        \State $J \leftarrow J + \nabla L_{\text{base}}$ \Comment{Eq.~\eqref{eq:loss_1}}
        
    \EndFor \label{alg2-line-attention-end}
    \State Form $f$ with updated $\mathbf{a, W}$ \Comment{Eq.~\eqref{eq:fx}}
    \State Compute $\mathcal{D}_S^{+}$, $\mathcal{D}_S^{-}$, $\bar{d}^{+}_{\mathcal{D}_S}$, $\bar{d}^{-}_{\mathcal{D}_S}$  \Comment{Eq.~\eqref{eq:centroids}} \label{alg2-line-attention-support-start}
    
    \State $L_{\text{ssl}} \leftarrow L_{\text{base}} + \phi L_{\text{support}}$\Comment{Eq.~\eqref{eq:loss_2}}
    \label{alg2-line-attention-support-end}
    
    \State $J \leftarrow J + \nabla L_{\text{ssl}}$ \Comment{Update $\mathbf{a}, \mathbf{W}, \mathbf{V}, \mathbf{b}$}

\EndLoop


    

\State Infer $\hat{\mathbf{y}}$ \Comment{Same as Line~\ref{alg1-line-reference-start-before}-~\ref{alg1-line-reference-end} of Algorithm~\ref{algo:method-un}}


\State \textbf{return} $\hat{\mathbf{y}}$, $\mathbf{a, W}$

\end{algorithmic}
\label{algo:method-ssl}
\end{algorithm}

\subsubsection{Hybrid Model}
\label{sec:method-hybrid}

We further propose a hybrid model that incorporates both the labeled support set as well as the unlabeled data in the target domain in the training process. It can be seen as the composition of \methodA and \methodB.
The loss function is as follows.
\begin{equation}
    L_{\text{hybrid}} = (1-\lambda) L_{\text{base}} + \lambda L_{\text{target}} + \phi L_{\text{support}}
    \label{eq:loss_hyb}
\end{equation}

This variant uses the loss $L_{\text{target}}$ defined in Equation~\eqref{eq:loss_un_target} and $L_{\text{support}}$ defined in Equation~\eqref{eq:loss_support}. We name this hybrid solution as \textbf{\methodC}. 
\dijin{
The algorithm is similar to Algo.~\ref{algo:method-ssl}, the main difference is to incorporate $L_{\text{un}}$ \ie, Equation~\eqref{eq:loss_un} into the training process (line~\ref{alg2-line-attention-start}-~\ref{alg2-line-attention-end}) to learn the parameters simultaneously. 
}

\subsection{Parameter Complexity}
\label{sec:method-analysis}

We measure the parameter complexity of \method in terms of the numbers of learnable parameters that comes from three parts: (i) per-feature non-linear affine operations that transform the word token embeddings to the latent feature vectors, (ii) the shared feature attention embedding function $f$, which includes learning $\mathbf{W}$ and $\mathbf{a}$, and (iii) the multilayer perceptron (MLP) $\Theta$ with 1 hidden layer for classification. For (i), there are totally $F$ features, each feature is associated with $\mathbf{V}^{H\times D}$ and $\mathbf{b}$, thus leading to $\mathcal{O}(FDH)$ learnable parameters. For (ii), as $\mathbf{W}^{H^{\prime}\times H}$ and $\mathbf{a}^{H^{\prime}}$ are shared across all features, there are totally $\mathcal{O}(HH^{\prime})$ parameters. 
The neural network $\Theta$ in (iii) takes the concatenated $FH^{\prime}$-dim features as input with one $H_{\text{hidden}}$-dim hidden layer. 
Therefore, \method has totally $\mathcal{O}(FDH + HH^{\prime} + FH^{\prime}H_{\text{hidden}})$ parameters to learn.
We discuss the setup values of $H$, $H^{\prime}$ and $H_{\text{hidden}}$ in the configuration of Section~\ref{sec:exp},
and empirically estimate the parameter number in Section~\ref{sec:support_set_analysis}.

\section{Experiments}
\label{sec:exp}

In this section we describe the experiments to evaluate properties and the performance of \method. Specifically, we aim to answer the following research questions:
\dijin{
\textbf{Q1.} Does \method effectively handle MEL with the data challenges (\textbf{C1}-\textbf{C3}) under the transfer learning paradigm?
\textbf{Q2.} How well does \method adapt feature importance in the target domain and how does it affect the linkage results?
\textbf{Q3.} Are generated feature attention values meaningful? \textbf{Q4.} How stable is \method in handling different data sources?
\textbf{Q5.} How does the size of support set $\mathcal{S}_U$ impact the performance of \method?
We conclude with the model justification (ablation study, limitation).
}

\begin{table}[b]
\centering
\caption{Data statistics and properties}
\label{table:data-stats}
\vspace{-0.3cm}
\resizebox{\columnwidth}{!}{\setlength{\tabcolsep}{4pt} \begin{tabular}{lrr C{0.7cm} C{0.7cm} C{0.6cm} }
\toprule
  \textbf{Data} & \# Records & Entity\_types & $|\mathcal{D}^{\ast}_S|$ & $|\mathcal{D}^{\ast}_T|$ & $|\mathcal{A}|$ \\ \hline
\texttt{Monitor} & 66,795 & Monitor & 5 & 24 & 13 \\
\texttt{Music-3K} & 3,070 & Artist, Album, Track & 3 & 7 & 9 \\
\texttt{Music-1M} & 1,723,426 & Artist, Album & 3 & 7 & 9 \\
\bottomrule
\end{tabular}
}
\vspace{-.2cm}
\end{table}

\subsection{Experimental Setup}
\noindent\textbf{Data} 
In accordance with (\textbf{Q1}-\textbf{Q5}), we use both the public benchmark dataset from the Data Integration to Knowledge Graphs (DI2KG) challenge~\cite{di2kg} and two real-world datasets in different scales from an online-sales company. Both datasets are in the tabular form and the entities are associated with descriptive textual features.
The data statistics and source info is given in Table~\ref{table:data-stats}.

\begin{itemize}
    
    \item \textbf{Music-1M} is a weakly labeled corpus crawled from $7$ public music websites. We name them \emph{website 1-7} for confidentiality. There are $2$ entity types: artists and albums. Entity pairs are labeled following the hyperlinks in pages, so there might be mixed-type errors such as matching an artist to her album.
        
    \item \textbf{Music-3K} is a manually labeled corpus containing the same data sources as \textbf{Music-1M}. It has three types: artist, album and tracks. The manual annotation is based on $9$ attributes such as the artist name and album title. Errors such as mixed-type matching are carefully corrected.
    
    \item \textbf{Monitor} contains monitor data from $24$ sales websites such as \emph{ebay.com} and \emph{shopmania.com}. We filter out attributes with $>60\%$ empty records, and get totally $13$ attributes such as product description, manufacturer info, condition status, etc.
\end{itemize}

Comparing with the public benchmark datasets~\cite{mudgal2018deep}, the above datasets are collected from larger ranges of real-world data sources with heterogeneous schemas. 
The attribute values in the above datasets are generally longer with diverse characters, which makes it harder to summarize the attribute representation. 
For example, for \texttt{Music-3K}, artist type, the averaged attribute length is $25.75$ word tokens, and for \texttt{Monitor}, the averaged attribute length is $11.73$ word tokens.
On the contrary, this number is $6.26$ and $5.21$ word tokens for the benchmark ``dirty'' and ``heterogeneous'' \texttt{Walmart-Amazon} dataset~\cite{fu2020hierarchical}, respectively.
In terms of the \texttt{Music} datasets, as the music works come from different countries, many entities are recorded with non-English characters \& phrases for attributes such as the title, album and artist names.
Unlike \texttt{Music-1M} that labels entity pairs through website hyperlinks, \texttt{Music-3K} further inspects whether the pair of music works indicate the same physical copy (\ie, ``Album''), or the same digital copy in formats such as remix or cover (\ie, ``Track'').
\dijin{
The \texttt{Monitor} dataset is highly imbalanced with more than $99\%$ entity pairs being unmatched. 
Additionally, less than 50\% entity pairs in this dataset have non-missing values for most attributes. Non-missing pairs of 5 attributes only exist in the target domain. These data challenges do not exist in the benchmark datasets~\cite{mudgal2018deep}, and we detail the analysis in Section~\ref{sec:public-data-analytics} of the supplementary material.
All the above issues make the datasets more challenging and closer to the real-world knowledge integration scenario.
}

\begin{table}[t!]
\vspace{-0.1cm}
\centering
\caption{Train, support and test statistics in the experiments.}
\label{table:data-split}
\vspace{-0.3cm}
\resizebox{\columnwidth}{!}{\setlength{\tabcolsep}{3pt} \begin{tabular}{lc C{1.6cm} C{2.cm} C{1.4cm} }
\toprule
  \textbf{Data} & \textbf{Entity\_type} & \textbf{Train} $|\mathcal{D}_S|$ & \textbf{Support} $|\mathcal{S}_U|$ & \textbf{Test} $|\mathcal{D}_T|$ \\ \hline
\multirow{3}{*}{\texttt{Music-3K}} 
   & Artist & 374 & 100 & 541 \\
   & Album  & 490 & 100 & 509 \\
   & Track  & 314 & 100 & 542 \\\hline
\multirow{2}{*}{\texttt{Music-1M}} 
   & Artist & 298\,566 & 100 & 541 \\
   & Album  & 697\,739 & 100 & 509 \\\hline
\texttt{Monitor} 
  & Monitor & 17\,766 & 100 & 1\,432 \\
\bottomrule
\end{tabular}
}
\vspace{-.cm}
\end{table}

\noindent\textbf{Baselines.}
The following baselines are used in this work.
\begin{itemize}[topsep=3px,partopsep=0px]
        \item \dijin{\textbf{TLER}~\cite{thirumuruganathan2018reuse} is a non-deep transfer learning framework that defines a standard feature space and reuses the seen data to train models for the new domain.
    }
    \item \textbf{DeepMatcher}~\cite{mudgal2018deep} is a deep learning framework that consists of 3 modules: attribute embedding, attribute similarity representation, and classification. The public implementation uses Fasttext to embed attribute words and uses attentative RNN to summarize attributes. We report results using the best-performing variant, DeepMatcher-hybrid.
        \item \textbf{EntityMatcher}~\cite{fu2020hierarchical} is a hierarchical deep framework for heterogeneous schema matching. It jointly matches entities at the level of token, attribute, and entity. The token-level matching strategy allows EntityMatcher to perform cross-attribute alignment. Fasttext is used to embed word tokens.
    \item \textbf{Ditto}~\cite{li2020deep} is an EL system that leverages fine-tuned, pre-trained Transformer-based language models (\ie, BERT, DistilBERT, or RoBERTa) with optimization including domain knowledge injection, text summarization, and data augmentation with difficult samples. 
        \item \textbf{CorDel}~\cite{wang2020cordel} adopts an alternative deep architecture to the widely-used ``twin architecture''. It compares and contrasts word tokens to filter out minor deviations between attribute values before embedding.
    CorDel also uses Fasttext, and it shows higher performance with reduced runtime compared to DeepMatcher. Out of the variants, CorDel-Attention is reported to perform the best on dirty EL datasets.
\end{itemize}
We consider these baselines since they are reported to achieve state-of-the-art EL performance and outperform methods such as Seq2SeqMatcher~\cite{nie2019deep} and DeepMatcher+~\cite{kasai2019low}. Most of them are particularly proposed to handle heterogeneous entity linkage.

\begin{figure*}[t!]
\vspace{-0.2cm}
\captionsetup[subfigure]{justification=centering}
  \centering
  \begin{subfigure}[t]{.92\textwidth}
      \centering
      \includegraphics[width=0.98\textwidth]{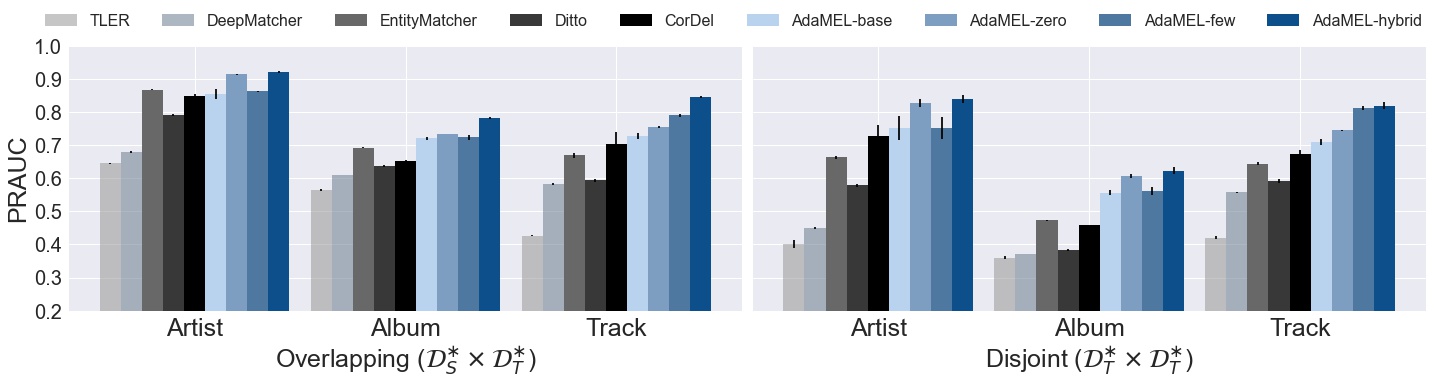}
      \vspace{-0.4cm}
      \caption{MEL performance on \texttt{Music-3K}}
  	\label{fig:mel-bar-a}
  \end{subfigure}
  \vspace{-0.6cm}
    \begin{minipage}[t]{0.6\textwidth}
    \vspace{-0.25cm}
    \centering
    \begin{subfigure}[t]{1.\textwidth}
        \includegraphics[width=1.\linewidth]{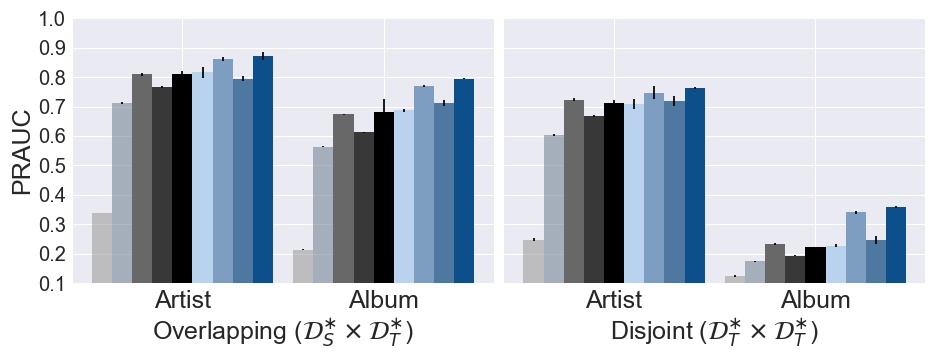}
        \vspace{-0.68cm}
        \caption{MEL performance on \texttt{Music-1M}}
        \label{fig:mel-bar-b}
    \end{subfigure}
  \end{minipage}  ~
  \begin{minipage}[t]{0.3\textwidth}
    \vspace{-0.2cm}
    \centering
    \begin{subfigure}[t]{1.\textwidth}
        \includegraphics[width=1.\linewidth]{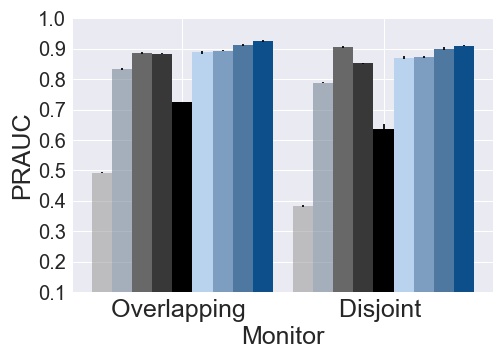}
        \vspace{-0.55cm}
        \caption{MEL performance on \texttt{Monitor}}
        \label{fig:mel-bar-c}
    \end{subfigure}
  \end{minipage}
  \vspace{0.4cm}
  \caption{MEL performance (PRAUC) comparison between \method variants and baselines. \method variants outperform baseline heterogeneous entity matching methods in almost all cases. Particularly, \methodC performs the best on all entity types and datasets.
  }
  
\vspace{-0.3cm}
\label{fig:mel-bar}
\end{figure*}

\noindent\textbf{Configuration.} 
In our experiments, we follow the original paper and fine-tune the baselines for optimal performance.
The statistics of training, support and testing data is given in Table~\ref{table:data-split}. Specifically, \texttt{Musci-1M} shares the same testing set as \texttt{Musci-3K}. \texttt{Monitor} adopts all positive and randomly selected $1000$ negative pairs to form the testing set (see Section~\ref{sec:public-code} of the appendix for more details).
For DeepMatcher, we use its hybrid variant (bi-directional RNN with attention) to summarize attributes with 2-layer highway neural network ((hidden dim$=300$)). 
The training epoches is set to $40$ with batch size $=16$.
For EntityMatcher, we use the full matching model that uses bi-GRU (hidden size$=150$) to embed attribute word sequences with cross-attribute token-level alignment.
The training epoch is set to $20$ with batch size $=16$.
For CorDel, we use the attention-based variant that learns the word importance within the same attribute to validate the effectiveness of our attribute-level attention module. Moreover, CorDel-Attention was shown to perform best on long textual attribute values, which matches the property of our input data.
All these 3 baselines use the pretrained FastText~\cite{joulin2017bag} to derive the 300-dimensional embeddings for word tokens in each attribute.
We set the cropping size $=20$ and sum the embeddings of word tokens as the feature embeddings for CorDel.
The training epoch is set to $20$ with learning rate $=10^{-4}$ and batch size $=16$.
For Ditto, we tested its optimization strategies and adopted the ``token span deletion'' for data augmentation, ``general'' domain knowledge and retaining high TF-IDF tokens to summarize the input sequences. We also tested all pretrained language models, \ie, bert, distilbert, and albert, and ended up using bert.
The training epoch is set to $40$ with batch size$=64$ and learning rate$=3\times 10^{-5}$.

To evaluate the effectiveness of our proposed framework, we configure \method with consistent setup as the baselines.
Specifically, we use the $300$-dim Fasttext to embed word tokens for fairness because $3$ of the $4$ baselines also use it, even though \method supports any word embedding techniques such as Bert embedding~\cite{kenton2019bert}. We set the cropping size$=20$ as CorDel.
The hyparameters of \method are given as follows:
the dimension of the projected embeddings per feature is $H=64$, the dimension of the hidden layer in $f$ is $H^{\prime}=256$, and the dimension of the hidden layer in $\Theta$ is $H_{\text{hidden}}=256$.
The activation $\sigma$ is set to be Relu. We set $\lambda=0.98$ and $\phi=1.0$ in Equation~\eqref{eq:loss_un}, ~\eqref{eq:loss_2} and ~\eqref{eq:loss_hyb} for \method variants unless otherwise addressed.
To train the \method, we use Adam optimizer~\cite{kingma2014adam} for $100$ epoches with learning rate $=10^{-4}$ and batch size $=16$.
We conduct all experiments 3 times and report the mean and std. We run these experiments on the Linux platform with 2.5GHz Intel Core i7, 256GB memory and 8 NVIDIA K80 GPUs.

\noindent\textbf{Evaluation Metric.} We evaluate the model performance using PRAUC as it measures the precision-recall relation globally and handles data imbalance.
We use the python Sklearn library to compute PRAUC based on the open-source implementation of all baselines.

\subsection{Transfer Learning for MEL}
\label{sec:exp1}

Our first experiment is to verify the effectiveness of \method variants on the task of MEL (\textbf{Q1}). We simulate two real-world scenarios: \textbf{(S1)} data in the target domain ($\mathcal{D}_T$) shares common data sources with the source domain ($\mathcal{D}_S$) (\ie, $(r, r^{\prime})_T\in \mathcal{D}^{\ast}_S\bigtimes \mathcal{D}^{\ast}_T$), and \textbf{(S2)} data sources in the target domain are disjointed from the source domain (\ie, $(r, r^{\prime})_T\in \mathcal{D}^{\ast}_T\bigtimes \mathcal{D}^{\ast}_T$).

\noindent\textbf{Setup.} 
For the \texttt{Music} data, we use three data sources (\ie, $\mathcal{D}^{\ast}_S = \{\text{\emph{website 1}, \emph{website 2}, \emph{website 3}}\}$) to train our model and test on all $7$ sources (overlapping scenario \textbf{S1}) or only the $4$ remaining sources (disjoint scenario \textbf{S2}) as the target domain $\mathcal{D}_T$. In either scenario, we collect $100$ samples ($50$ positive and $50$ negative) from the corresponding $\mathcal{D}_T$ as support set $\mathcal{S}_U$.
For the public \texttt{Monitor} data, we use entity pairs from 5 sources (\ie, $\mathcal{D}^{\ast}_S=\{$\emph{ebay.com}, \emph{catalog.com}, \emph{best-deal-items.com}, \emph{cleverboxes.com}, \emph{ca.pcpartpicker.com}$\}$) to train the models. We use data in all 24 sources as $\mathcal{D}_T$ for \textbf{S1}, and the rest 19 data sources for \textbf{S2}, respectively. $100$ samples are collected as $\mathcal{S}_U$ in the same way as \texttt{Music}.
\dijin{We also randomly picked different sources to form $\mathcal{D}_S$ and $\mathcal{D}_T$ to eliminate the randomness, and found similar patterns in the results.
}

\noindent\textbf{Results.} 
We report the results in Figure~\ref{fig:mel-bar} with complete numerical results in Table~\ref{table:exp1-a} and ~\ref{table:exp1-b} of Section~\ref{sec:complete-result} in the supplementary material.
Our first observation is that all \method variants tend to outperform the baseline methods and our base model without adaptation, \method-base. The heterogeneous entity matching baselines do not perform well on these datasets under the supervision of labeled data only. 
This is likely because of the long and noisy word sequences in the data and the difference in attribute value distribution across data sources that is unseen during model training. 
\method highlights the impact of important features, and only represent the sequences by summing the token embeddings.
This confirms our conjecture that learning the attribute-level attention as the transferable knowledge is more effective in handling the MEL task than refining the word-level sequence representation.
Also, we observe that out of all variants, \methodC achieves the best performance in all cases with $0.64\%\sim 5.50\%$ improvement in PRAUC than the second-best (\methodA in most cases), which demonstrates its effectiveness in integrating both the labeled support set $\mathcal{S}_U$ and unlabeled info from the target domain $\mathcal{D}_T$. 
\methodA performs better than \methodB on the ``Artist'' and ``Album'' type, while \methodB performs better on the ``Track'' type. This is likely due to the fact that the track records are more diverse than the other types as the digital-format music tracks can be remixed or covered by other artists. Thus, the high-quality labeled samples from $\mathcal{S}_U$ is of higher value. 
\dijin{
On the contrary, since the records are more consistent for the ``Artist'' and ``Album'' type, incorporating more records in $\mathcal{D}_T$ leads to higher improvement in MEL performance.
Note Figure~\ref{fig:mel-bar-b} shows that \methodB performs slightly worse than \method-base because the labeled samples from $\mathcal{S}_U$ only overfits to the trained model on the source domain, that deviates the actual feature importance for the massive unlabeled samples in $\mathcal{D}_T$.
}
To summarize, the improvement of \methodA, -\textsc{few} and -\textsc{hyb} over the baselines indicates the effectiveness of domain adaptation in incorporating data in $\mathcal{D}_T$.

Overall, \method variants achieve better performance on the overlapping scenario (\textbf{S1}) than the disjoint scenario (\textbf{S2}). This is as expected as the disjoint scenario represents an extreme case where data sources in $\mathcal{D}_T$ are less likely or even entirely not used in training the model if the support set is unavailable.
Besides, the performance of all approaches running on \texttt{Music-1M} is lower than \texttt{Music-3K}. 
The main reason is that the data is weakly labeled as it simply follows the hyperlinks from the websites, and does not distinguish the actual media of the music work (\ie, the physical or digital copy). As the models are tested on the same well-labeled set, training on \texttt{Music-1M} could be vulnerable to cases such as mixed-type matching.
Nevertheless, we observe that \method still achieves promising results in both hard cases of transfer learning for MEL, \ie, unseen data sources in the target domain and training on weakly labeled data, which further demonstrates the advantage of \method.
Table~\ref{table:exp1-b} gives the result on \texttt{Monitor}. Similarly, \method variants tend to outperform the baselines and \methodC performs the best with at least $0.51\%$ improvement in PRAUC over the second best, EntityMatcher. On average, \methodC outperforms the baseline by $9.39\%$ improvement in the overlapping scenario and $11.55\%$ improvement in the disjoint scenario. 
These results also validates our findings above.

\begin{figure}[t]
\vspace{-.5cm}
\captionsetup[subfigure]{justification=centering}
  \centering
  \begin{subfigure}[t]{0.25\textwidth}
      \centering
      \includegraphics[width=0.98\textwidth]{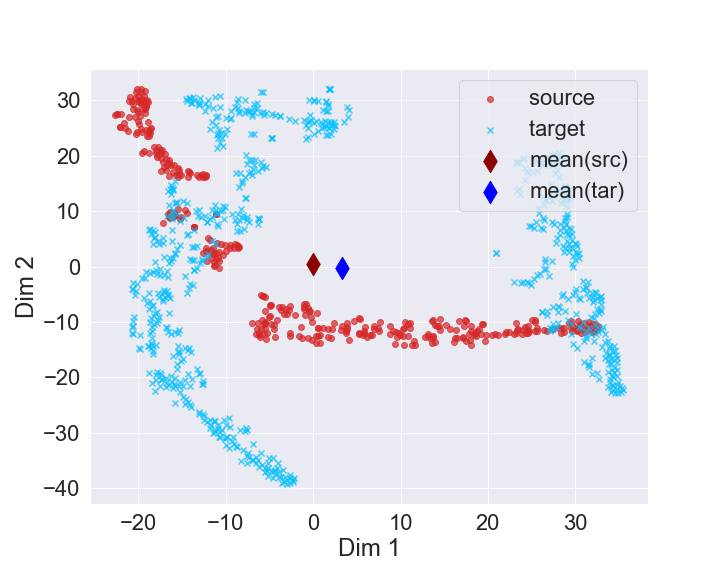}
      \vspace{-0.2cm}
      \caption{\methodA with no adaptation ($\lambda = 0$)}
  	\label{fig:adaptation_compare_a}
  \end{subfigure}
  ~
   \begin{subfigure}[t]{0.25\textwidth}
      \centering
      \includegraphics[width=0.98\textwidth]{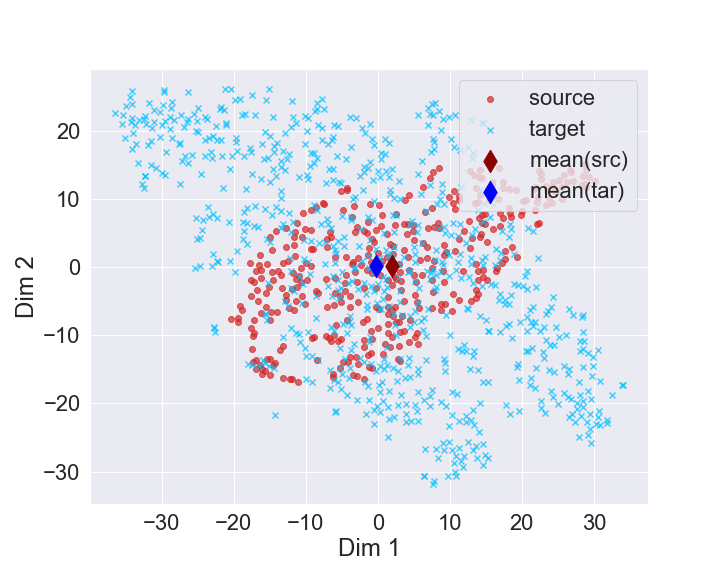}
      \vspace{-0.2cm}
      \caption{\methodA with adaptation ($\lambda = 0.98$)}
      \label{fig:adaptation_compare_b}
  \end{subfigure}
  
   \begin{subfigure}[t]{0.25\textwidth}
      \centering
      \vspace{-0.1cm}
      \includegraphics[width=0.98\textwidth]{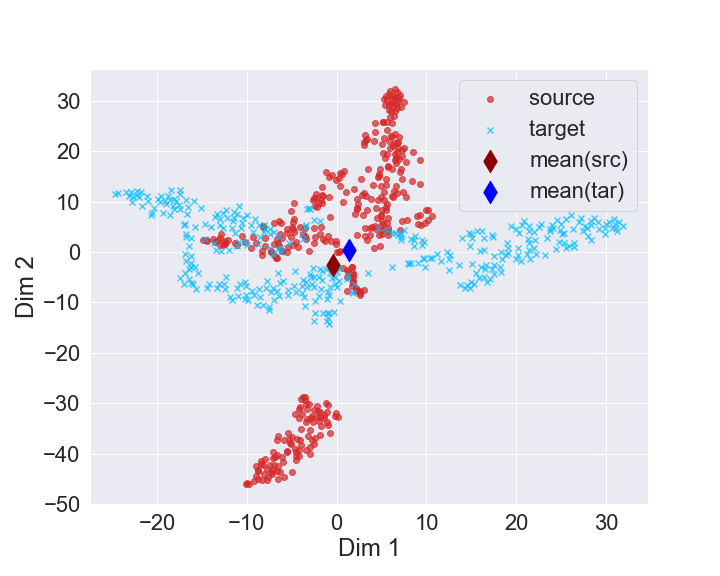}
          \caption{\methodC with no adaptation ($\lambda = 0$)}
      \label{fig:adaptation_compare_c}
  \end{subfigure}
  ~
   \begin{subfigure}[t]{0.25\textwidth}
      \centering
      \vspace{-0.1cm}
      \includegraphics[width=0.98\textwidth]{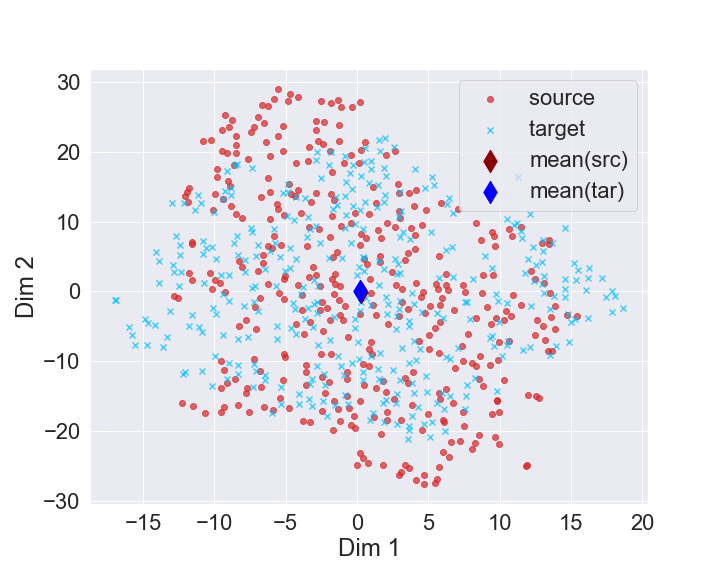}
          \caption{\methodC with adaptation ($\lambda = 0.98$)}
      \label{fig:adaptation_compare_d}
  \end{subfigure}
  \vspace{-0.3cm}
  \caption{Source and target domain feature attention vectors are better aligned with high value of $\lambda$ for both \methodB and \methodC (visualized with TSNE, dim=2).
  }
  
\label{fig:adaptation_compare}
\end{figure}

\subsection{Effectiveness of Adaptation}
\label{sec:exp2}
Adaptation is the key component to our proposed method.
To evaluate how well \method learns feature importance adapted to the target domain (\textbf{Q2}), we study the effectiveness of $\lambda$ adopted in \methodA and \methodC as it controls the weight of adapting to unlabeled data in training (larger $\lambda$ leads to more adaptation).

\noindent\textbf{Setup.} We run both variants of \method on the \texttt{Music-3K} dataset and report the performance on MEL.
As discussed in Section~\ref{sec:method-zsl}, records from both the source and target domains are projected into the same space using the shared attention embedding function, and \method attempts to adapt the model to match these feature importance distribution.
Intuitively, with sufficient adaptation, feature importance vectors from both domains should align well, and further benefit the linkage task.
To validate this conjecture, we visualize the learned feature attention vectors using \methodA and \methodC with different values of $\lambda$ by projecting them into 2-dimensional space using TSNE~\cite{maaten2008visualizing}.
We also study the linkage performance of \methodA and \methodC with different $\lambda$ values on the ``artist'' and ``album'' type of the \texttt{Music-3K} dataset.

\begin{figure}[tp]
\vspace{-.2cm}
\captionsetup[subfigure]{justification=centering}
  \centering
  \begin{subfigure}[t]{0.245\textwidth}
      \centering
      \includegraphics[width=0.98\textwidth]{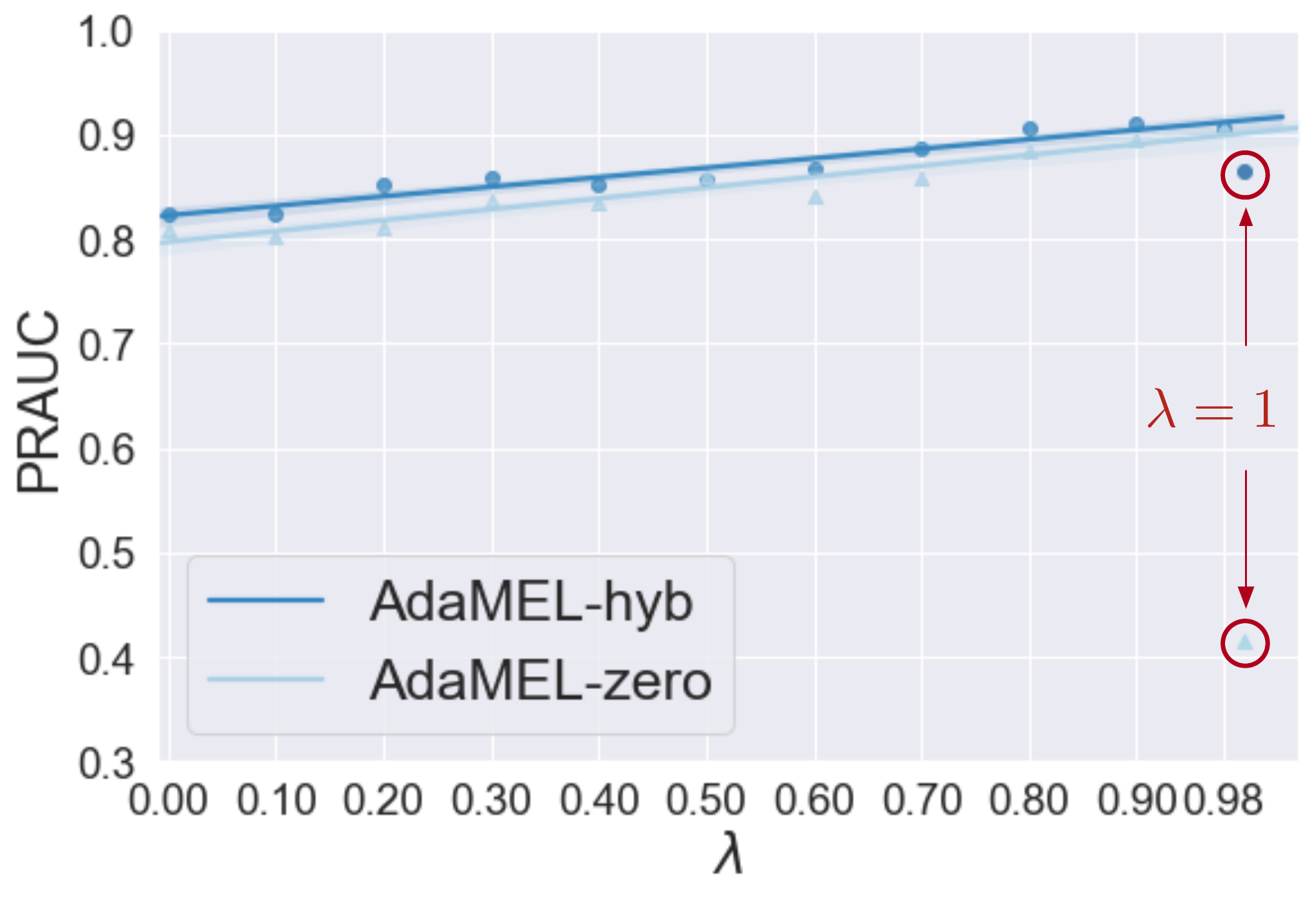}
          \caption{\method performance change on \texttt{Music-3K}, artist type}
  	\label{fig:adaptation_range_a}
  \end{subfigure}
  ~
   \begin{subfigure}[t]{0.245\textwidth}
      \centering
      \includegraphics[width=0.98\textwidth]{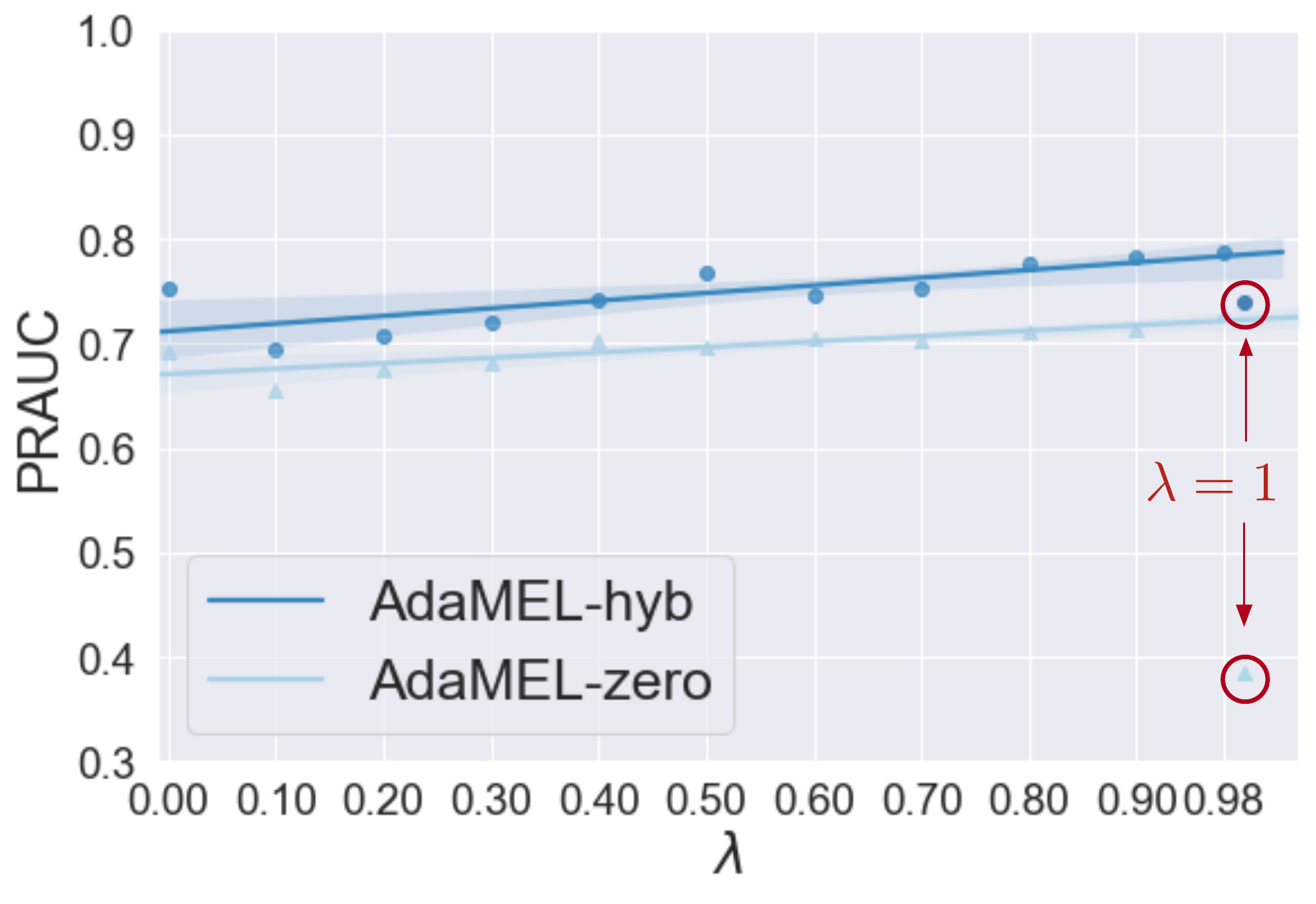}
          \caption{\method performance change on \texttt{Music-3K}, album type}
      \label{fig:adaptation_range_b}
  \end{subfigure}
  \vspace{-0.3cm}
  \caption{\methodA and \methodC performance improve with increasing $\lambda$ from $0$ to $0.98$ (fitted with linear regression). The performance drops when $\lambda=1$ as no labeled data in $\mathcal{D}_S$ is used.
  }
\vspace{-.2cm}
\label{fig:adaptation_range}
\end{figure}

\noindent\textbf{Results.} In Figure~\ref{fig:adaptation_compare}, we observe that for both variants, feature attention vectors from $\mathcal{D}_S$ and $\mathcal{D}_T$ align better when $\lambda = 0.98$ than $\lambda = 0$, which confirms the effectiveness of adaptation.
In addition, we observe that comparing with \methodA (Figure~\ref{fig:adaptation_compare_b}), \methodC (Figure~\ref{fig:adaptation_compare_d}) generates better adapted results as the projected records from $\mathcal{D}_S$ and $\mathcal{D}_T$ are almost indistinguishable, which is as expected as the labeled support set is leveraged. 

To evaluate the impact of adaptation to the linkage results,
in Figure~\ref{fig:adaptation_range} we show the performance of our variants with different $\lambda$ values.
We observe that as $\lambda$ approaches (but not equals) to $1$, the general performance in terms of PRAUC improves for both \methodA ($0.8014$ - $0.9091$) and \methodC ($0.8242$ - $0.9201$), which again demonstrates the effectiveness of adaptation.
It is worth noting that when $\lambda=1$, both \methodA and \methodC perform worse without giving meaningful results. 
This is because at this point, \methodA is trained without supervision of the labeling in $\mathcal{D}_S$, and the only term in the loss function is the regularization. \methodC is better as labeling in $\mathcal{S}_U$ is still used, but the overall performance deteriorates due to the lack of labeling from $\mathcal{D}_S$.
As a result, the parameters trained (\ie, $\mathbf{a, W}$) would tend to only ``match'' the feature distribution between $\mathcal{D}_S$ and $\mathcal{D}_T$ that are not tailored to classification.

\subsection{Attention Analysis}
\label{sec:exp4}

\noindent\textbf{Setup.} To testify whether \method learns meaningful feature attention values (\textbf{Q3}), 
we showcase the learned feature importance through the attention scores produced by \method on two datasets: \texttt{Music-3K} and \texttt{Monitor}. 
We only report the artist type and omit the other two types for brevity.
\methodC is configured with the best performance ($\lambda=0.98, \phi=1.0$) in the previous experiments.

\noindent\textbf{Result.} For the \texttt{Monitor} dataset in Table~\ref{table:attention_analysis-a}, we observe the long ``tail distribution'' of feature importance, \ie, the most important feature is ``Page\_title\_shared'' with significantly high scores, while the other features are with roughly the same low scores. 
On the other hand, we observe the more uniform distribution for the artist type in \texttt{Music-3K} dataset, which makes sense as all top features are related to the artist names. 
The learned attention scores on both datasets imply that the task of MEL could be addressed with some of the most remarkable features (importance inequality).

\begin{table}[tp]
\centering
\vspace{-.2cm}
\setlength{\tabcolsep}{5pt} \caption{\method learned importance of top-5 features for \texttt{Monitor} and \texttt{Music-3K}, artist type. }
\vspace{-0.3cm}
\resizebox{.96\columnwidth}{!}{
\begin{tabular}{@{}l|c|l|c}
\toprule
\multicolumn{2}{c|}{\texttt{Monitor}} & \multicolumn{2}{c}{\texttt{Music-3K}, artist} \\ 
\midrule
Feature & Score & Feature & Score \\ \hline
Page\_title\_shared      & 0.1635 & Main\_performer\_shared    & 0.0739  \\
Page\_title\_unique      & 0.0595 & Name\_unique    & 0.0697 \\
Source\_shared           & 0.0535 & Name\_shared        & 0.0628  \\
Manufacturer\_unique     & 0.0473 & Source\_unique    & 0.0597  \\
Manufacturer\_shared     & 0.0416 & Name\_Native\_Language\_shared  & 0.0583  \\
\bottomrule
\end{tabular}
}
\label{table:attention_analysis-a}
\vspace{-0.2cm}
\end{table}

We further run \methodC on these selected important features only and compare the performance with the result using the other features, as well as all the features. For \texttt{Monitor}, we use 3 attributes (\ie, ``Page\_title'', ``Source'' and ``Manufacturer''). 
For the artist type of \texttt{Music-3K}, we use the 3 name-related attributes (\ie, ``Main\_performer'', ``Name'', Name\_Native\_Language), and ``Source''.
Similarly, for the other two types, we use their corresponding top important attributes, and report the results in Table~\ref{table:attention_analysis-b}. 
We observe that by using the selected important features only, \method is capable of achieving comparable and even slightly better performance than using all features with $2.21\%$, $0.87\%$ and $2.92\%$ improvement in PRAUC on \texttt{Monitor}, \texttt{Music-3K} (artist) and \texttt{Music-3K} (album), respectively.
For \texttt{Music-3K} (track), using the top attributes only performs slightly worse than using all attributes, which is likely due to the diversity of track records.
Nevertheless, these experimental results show that model training can further benefit from feature importance as using all the possible attributes could introduce irrelevant or noisy input to the model (\eg, using album-related features when inferring the artist type).
Also, they shows the effectiveness of the feature attention module of \method in learning reasonable feature importance.

\begin{table}[hp]
\centering
\vspace{-0.2cm}
\setlength{\tabcolsep}{3pt} \caption{Performance (PRAUC) comparison using the selected important features vs. the other features and all features. Numbers in the parenthesis denote the counts of features.}
\vspace{-0.3cm}
\resizebox{\columnwidth}{!}{
\begin{tabular}{@{}l|c|c|cH}
\toprule
Dataset & Top Attributes (\#) & Other Attributes (\#) & All Attributes (\#) & Selected vs. All \\ 
\midrule
\texttt{Monitor}                & \textbf{0.9479 $\pm$ 0.0007} (3) & 0.4276 $\pm$ 0.0015 (10) & 0.9258 $\pm$ 0.0025 (13) & + 2.21 \%  \\
\texttt{Music-3K}, artist     & \textbf{0.9298 $\pm$ 0.0036} (4) & 0.7966 $\pm$ 0.0005 (5) & 0.9211 $\pm$ 0.0040 (9) & + 0.87 \%  \\
\texttt{Music-3K}, album     & \textbf{0.8125 $\pm$ 0.0011} (4) & 0.4692 $\pm$ 0.0009 (5) & 0.7833 $\pm$ 0.0031 (9) & + 0.87 \%  \\
\texttt{Music-3K}, track     & 0.8398 $\pm$ 0.0004 (3) & 0.7026 $\pm$ 0.0006 (6) & \textbf{0.8454 $\pm$ 0.0040} (9) & + 0.87 \%  \\
\bottomrule
\end{tabular}
}
\label{table:attention_analysis-b}
\vspace{-0.3cm}
\end{table}

\subsection{Data Sources Analysis}
\label{sec:data_source_analysis}
\dijin{
In this experiment we simulate the real-world knowledge integration, where new data sources often arrive one by one incrementally (such as in batches from neighboring data sources), and testify the stability of \method in handling the various data sources under this scenario (\textbf{Q4}) .
}

\noindent\textbf{Setup.} We use the public \texttt{Monitor} dataset and compare \methodC with the optimal configuration ($\lambda=0.98, \phi=1.0$ as shown in Section~\ref{sec:exp1} and Section~\ref{sec:exp2}) with the best-performing baseline approach, EntityMatcher, and the fastest baseline approach, CorDel-Attention.
In this experiment, we use $1500$ entity pairs from the same 5 data sources as mentioned in Section~\ref{sec:exp1} to train the models (\ie, $\mathcal{D}_S^{\ast}=\{$\emph{ebay.com}, \emph{catalog.com}, \emph{best-deal-items.com}, \emph{cleverboxes.com}, \emph{pcpartpicker.com}$\}$).
To test the performance on MEL, we first randomly select $200$ entity pairs from each of $7$ data sources (the same 5 data sources as $\mathcal{D}_S^{\ast}$ and 2 unseen ones, \ie, $\mathcal{D}_T^{\ast}=\mathcal{D}_{S}^{\ast} \cup \{$\emph{yikus.com}, \emph{getprice.com}$\}$) and form totally $1400$ pairs to create the target domain.
Then, we incrementally add up to $200$ entity pairs from 2 new sources ($\Delta\mathcal{D}_T^{\ast}$) to $\mathcal{D}^{\ast}_T$, such that $\mathcal{D}^{\ast}_T=\mathcal{D}^{\ast}_T\cup\Delta\mathcal{D}^{\ast}_T$. Each of the newly added pairs $\{(r, r^{\prime})\}$ contains at least one record from $\Delta\mathcal{D}_T$ to ensure new data sources are introduced to the target domain.
\dijin{
As \methodC requires a small set of labeled entity pairs from $\mathcal{D}_T^{\ast}$, we randomly select $100$ labeled samples from all data sources ($\mathcal{D}_S^{\ast}\cup\mathcal{D}_T^{\ast}$). This small set simulates the on-the-fly manual labeling in the real-world, and we fix it throughout each run of the experiment to ensure the impact of $\mathcal{S}_U$ is consistent.
We also record the average runtime over all runs as an empirical study of the model efficiency.
}

\begin{figure}[tp]
\vspace{-0.4cm}
  \begin{minipage}[t]{0.68\columnwidth}
    \vspace{0pt}
    \centering
    \includegraphics[width=1.\linewidth]{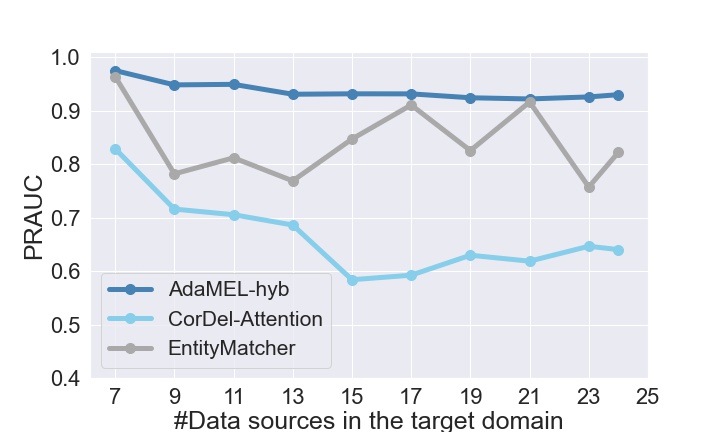}
    \label{fig:source_analysis}
  \end{minipage}  \hspace{-0.2cm}
  \begin{minipage}[t]{0.3\columnwidth}
    \vspace{1.1cm}
    \centering
        \resizebox{1.\columnwidth}{!}{
        \begin{tabular}{@{}l|l}
            \toprule
            Method & Runtime (s) \\ \hline
            Hybrid & 319.20 $\pm$ 7.20 \\
            CorDel & 906.19 $\pm$ 46.35 \\
            E-Matcher & 2500.43 $\pm$ 17.56 \\
            \bottomrule
            \end{tabular}
            }
  \end{minipage}
  \vspace{-.5cm}
  \caption{\methodC performs more stably ($0.9750 \sim 0.9219$ in PRAUC) as \#data sources increases in $\mathcal{D}_T$ with less runtime.}   \label{fig:source_analysis}
  \vspace{-0.2cm}
\end{figure}

\noindent\textbf{Results.} We report the performance of \methodC and the two baselines on MEL in Figure~\ref{fig:source_analysis}, as well as their empirical runtime.  As shown in the figure, \methodC is more stable than both EntityMatcher and CorDel-Attention with significantly higher performance in handling the incrementally incoming data sources. 
This is due to the fact that \methodC continuously updates parameters in the attention embedding function $f$ to adapt to new data sources in $\mathcal{D}_T$.
Comparing with CorDel-Attention, EntityMatcher performs better and could occasionally compete with \methodC under some scenarios ($|\mathcal{D}^{\ast}_T| = 17, 21$), but it is not stable as the performance fluctuates.
Moreover, based on the table in Figure~\ref{fig:source_analysis}, \methodC takes much less time to train than CorDel-Attention and EntityMatcher.
The empirical runtime comparison corresponds to our analysis in Section~\ref{sec:method-analysis} as \methodC does not require sophisticated operations on word-level embeddings and thus having relatively less parameters to train.
In practice, the number of parameters to train for \methodC is $\sim2\,219\,520$, which is much less than the number given by EntityMatcher: $\sim123\,119\,104$.
These findings demonstrate the capability of \method in consistently handing MEL with a variety of incoming data sources, while being more robust.
In addition, they strengthen our claim that finding important features as the transferable knowledge in MEL could benefit the model performance with reduced computational complexity.

\subsection{Effectiveness of Support Set}
\label{sec:support_set_analysis}

\noindent\textbf{Setup.} To better understand the effectiveness of the labeled support set (\textbf{Q5}), we perform the sensitivity analysis with incrementally increasing numbers of labeled samples in the support set $\mathcal{S}_U$. Following Section~\ref{sec:exp1}, we randomly select 200 additional samples from $\mathcal{D}_T$ of the public \texttt{Monitor} dataset and create the support set with totally 300 labeled samples. 
We run two \method variants that leverage the support set, \methodB ($\phi=1.0$) and \methodC ($\lambda=0.98, \phi=1.0$) in this experiment with $|\mathcal{S}_U|$ ranging from $1$ to $300$ with step size $=20$ (specifically, we ``zoom in'' the smaller values and have $|\mathcal{S}_U|=\{1, 5, 10, 20, 40, 60, \cdots, 300\}$). 
In each run, the samples in $\mathcal{S}_U$ are randomly selected.

\begin{figure}[tp]
\vspace{-.3cm}
\centering
\includegraphics[width=0.88\linewidth]{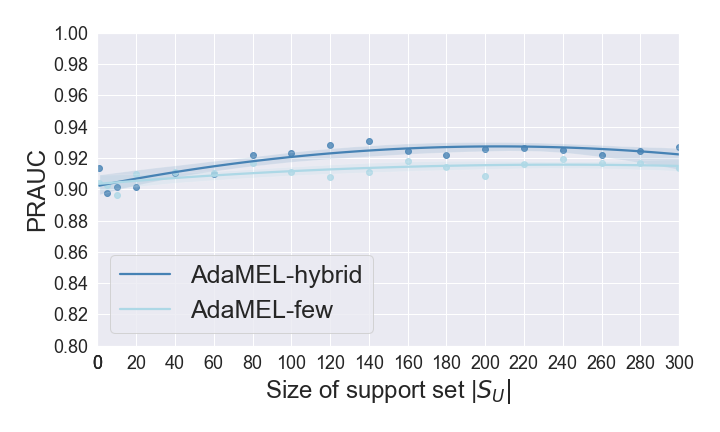}
\vspace{-.4cm}
\caption{
Sensitivity analysis of the size of support set $|\mathcal{S}_U|$ fitted with order-2 polynomial regression on \methodB and \methodC. As more labeled samples are included in $\mathcal{S}_U$, the model performance (PRAUC) increases initially and then flattens out. 
}
\label{fig:support_set}
\end{figure}

\noindent\textbf{Result.} The experimental result is shown in Figure~\ref{fig:support_set}. 
Our first observation is that at the initial stage of the experiment, the performance of both \methodB and \methodC improves as the number of used labeled samples from $\mathcal{S}_U$ increases. 
Particularly, we observe $\sim 1\%$ performance improvement from $|\mathcal{S}_U| = 1$ to $|\mathcal{S}_U| = 140$ for \methodB and $2\%\sim3\%$ improvement for \methodC.
\dijin{
This overall performance improvement is as expected since an increasing amounts of labeled samples from $\mathcal{D}_T$ are used to supervise the learning process. 
In the late stage ($|\mathcal{S}_U| > 140$), we observe that the performance fluctuates in each run and the overall performance saturates. 
This indicates that the feature importance learned by \method has sufficiently adapted and does not significantly change as more labeled data are collected in $\mathcal{S}_U$.
Moreover, comparing with \methodB, \methodC performs similarly when the size of support set is small ($|\mathcal{S}_U| \le 60$), and it consistently outperforms when $|\mathcal{S}_U| > 60$. 
This is likely due to the bias of feature importance brought by particular labeled samples selected when $|\mathcal{S}_U|$ is small. 
}
When $\mathcal{S}_U$ contains more samples, the learned feature importance becomes stable and sufficiently adapted to $\mathcal{S}_U$, and the outperformance given by \methodC over \methodB comes from the unlabeled samples from $\mathcal{D}_T$.
As a rule of thumb, Figure~\ref{fig:support_set} indicates that a small support set with $|\mathcal{S}_U|=100 \sim 200$ labeled samples from $\mathcal{D}_T$ is beneficial to learn feature importance and to improve the MEL performance of \method. 
Too few samples would incur bias to the trained model, while too many samples would be expensive to obtain in practice, and does not necessarily help improve the model.

\color{black}
\subsection{Model Justification}
In this section we run experiments to justify the design choices of \method and its limitation.

\subsubsection{Ablation Study}
\label{sec:ablation_study}

We perform the ablation study of \method that uses the shared and unique contrastive features, as well as using both of them as the default setting. Table~\ref{table:ablation} shows that including the shared and unique attribute values capture different perspectives of the data and thus enriches the feature space. Including both achieves the highest performance with $0.41\%-6.72\%$ improvement over using one feature.

\begin{table}[t]
\centering
\setlength{\tabcolsep}{5pt} \caption{\dijin{Ablation study: \method contrastive features on \texttt{Music-3K}, artist and album type. \methodA and -\textsc{few} perform similarly.}}
\vspace{-0.3cm}
\resizebox{\columnwidth}{!}{
\begin{tabular}{@{}l|c|c|c|c}
\toprule
Dataset & Method & Shared & Unique & Shared \& Unique \\ 
\midrule
\multirow{2}{*}{\shortstack[l]{\texttt{Music-3K},\\artist}} & 
  \method-base & 0.7868 $\pm$ 0.0045 & 0.7170 $\pm$ 0.0132 & \textbf{0.8545 $\pm$ 0.0143}   \\
& \methodC     & 0.8539 $\pm$ 0.0026 & 0.8069 $\pm$ 0.0112 & \textbf{0.9211 $\pm$ 0.0040}  \\
\midrule
\multirow{2}{*}{\shortstack[l]{\texttt{Music-3K},\\album}} & 
  \method-base & 0.7163 $\pm$ 0.0048 & 0.5520 $\pm$ 0.0044 & \textbf{0.7204 $\pm$ 0.0033}  \\
& \methodC     & 0.7504 $\pm$ 0.0059 & 0.5879 $\pm$ 0.0028 & \textbf{0.7833 $\pm$ 0.0031}  \\
\bottomrule
\end{tabular}
}
\label{table:ablation}
\vspace{-0.1cm}
\end{table}

\subsubsection{Performance on Single Domain}
\label{sec:unadapted_exp}

Here we compare \methodA and \textsc{-hyb} with DeepMatcher on the benchmark datasets to justify their performance on well-labeled data from the same seen domain without the 3 challenges (\textbf{C1} - \textbf{C3}).
From Table~\ref{table:benchmark}, we observe that \methodA does not perform as well as DeepMatcher on these benchmark datasets of one single domain. This shows the limitation of \method in handling data with no missing values or schema difference. The reason is likely due to the simplicity of \method architecture, as it aims to learn the data-source-level feature importance instead of improving the token-level embeddings as DeepMatcher or its variants. In the real-world knowledge integration process where data distributions are highly heterogeneous, transferring these token-level contextualized embeddings brings extra computation and does not always generalize well, as shown in Section~\ref{sec:exp1}. Nevertheless, even though \method is designed to handle data challenges in practice (\textbf{C1}-\textbf{C3}), we observe that \methodC performs comparably as DeepMatcher with reduced model complexity, which shows its effectiveness of adaptation.

\begin{table}[ht]
\centering
\vspace{-0.1cm}
\setlength{\tabcolsep}{3pt} \caption{\dijin{Entity linkage performance (F1) of DeepMatcher, \methodA and -\textsc{hyb} on the benchmark datasets, single domain scenario. \methodC performs comparably as DeepMatcher.}}
\vspace{-0.3cm}
\resizebox{\columnwidth}{!}{
\begin{tabular}{@{}llcccc}
\toprule
Type & Datasets & Domain & DeepMatcher & \methodA & \methodC \\ 
\midrule

\multirow{7}{*}{Structured} 
& Amazon-Google  & Software    & 69.3 & 60.2 & 65.1   \\
& Beer           & Product     & 78.8 & 78.6 & 82.8   \\
& DBLP-ACM       & Citation    & 98.4 & 98.7 & 98.9   \\
& DBLP-Google    & Citation    & 94.7 & 93.1  & 93.5 \\
& Fodors-Zagats  & Restaurant  & 100  & 90.0 & 99.8 \\
& iTunes-Amazon  & Music       & 91.2 & 91.2 & 98.7 \\
& Walmart-Amazon & Electronics & 71.9 & 57.8 & 66.7 \\
\midrule
\multirow{4}{*}{Dirty} 
& DBLP-ACM       & Citation    & 98.1 & 95.7 & 97.7 \\
& DBLP-Google    & Citation    & 93.8 & 89.7 & 91.5 \\
& iTunes-Amazon  & Music       & 79.4 & 79.3 & 80.7\\
& Walmart-Amazon & Electronics & 53.8 & 48.2 & 52.2 \\
\bottomrule
\end{tabular}
}
\label{table:benchmark}
\vspace{-0.3cm}
\end{table}

\color{black}

\section{Conclusion}
\label{sec:conclusion}
In this work, we tackle the problem of multi-source entity linkage (MEL) and described a deep learning solution based on domain adaptation, \method.
\method highlights the impact of important attributes in MEL and automatically learns feature importance that adapts to the both seen and unseen data sources as the generic transferable knowledge.
We also propose a series of \method variants to handle different real-world learning scenarios, depending on the availability of labeled entity pairs from the target domain.
Comparing to heterogeneous schema matching baselines, \method is able to handle hard transfer learning cases such as unseen data sources in the target domain and training on weakly-labeled data, while achieving on average $8.21\%$ improvement than the baselines based on supervised learning in PRAUC score for the multi-source entity linkage task.
In addition, our experiments demonstrate the
effectiveness of \method in adaptation, provide the analysis on the learned feature attention, and study the impact of different data sources as well as the size of the support set.
Future directions include combining our work with advanced NLP techniques for sequence representation in attribute summarization to further improve the model performance in MEL, 
and extending our framework to handle data sources in different languages.

\clearpage
\balance
\bibliographystyle{ACM-Reference-Format}
\bibliography{main}

\clearpage

\appendix
\thispagestyle{empty}
 
\section{Supplementary Materials}
\label{sec:supp-materials}

\subsection{Public Data Processing}
\label{sec:public-code}
In this section we detail the processing of the public dataset, \texttt{Monitor} that is used in the experiment.
We follow the \emph{'monitor\_label.csv'} file ($1\,073$ positive pairs and $110\,082$ negative pairs) from the DI2KG to create the labeled entity pairs that fall into the 24 data sources we are interested in. The resultant dataset, \texttt{Monitor} has $734$ positive and $66\,061$ negative pairs.
The source domain $\mathcal{D}_S^{\ast}$ contains 5 data sources (\ie, $\mathcal{D}^{\ast}_S=\{$\emph{ebay.com}, \emph{catalog.com}, \emph{best-deal-items.com}, \emph{cleverboxes.com}, \emph{ca.pcpartpicker.com}$\}$), which contains $302$ positive pairs.
Thus, the test data includes all the remaining positive $432$ pairs and randomly-selected $1,000$ negative pairs.

We provide the code and the splitted public \texttt{Monitor} data in the following in the repo \url{https://github.com/DerekDiJin/AdaMEL-supplementary}.

\begin{figure}[hb]
\vspace{.2cm}
\centering
\includegraphics[width=0.92\linewidth]{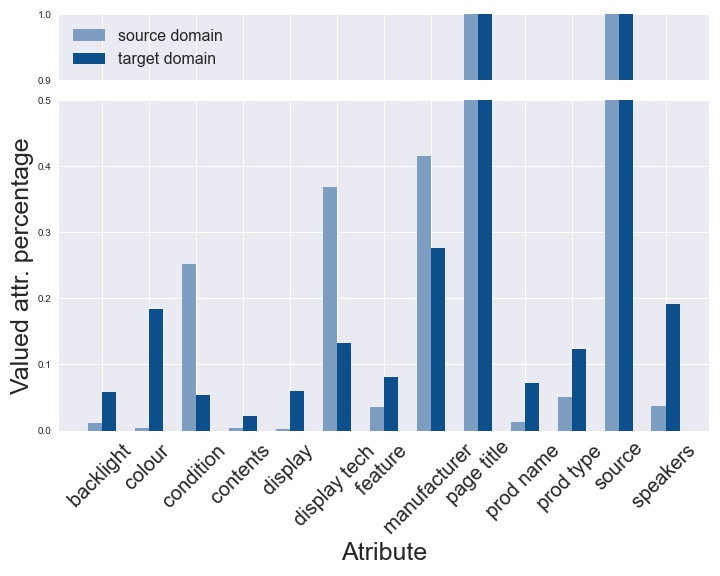}
\vspace{-.4cm}
\caption{
\texttt{Monitor}: the challenges of missing attribute values (\textbf{C1}) and new attributes (\textbf{C2}) between $\mathcal{D}_S$ and $\mathcal{D}_T$ is shown with the percentages of entity pairs without missing values per attribute (\ie, nonempty for both entities). For most attributes, the majority of entity pairs have at least 1 entity with missing values. 5 out of 13 attributes have non-missing entity pairs only in the target domain (2 non-missing attributes are ``page title'' and ``source''). For the remaining 6 attributes, the percentage is also significantly different between the source and target domain.
}
\label{fig:data_stats}
\end{figure}

\dijin{
\subsection{Analysis of Public Data}
\label{sec:public-data-analytics}
To illustrate the data challenges (\textbf{C1}-\textbf{C3}), we provide detailed analysis of the \texttt{Monitor} data. We first study the difference of the source and target domain in terms of the attributes, \ie, missing attribute values (\textbf{C1}) and new attributes (\textbf{C2}).
As the attributes associated with entity pairs, we plot the percentage of pairs without missing values per attribute, \ie, $(r[A], r^{\prime}[A])$ where $r[A] \ne \empty, r^{\prime}[A] \ne \empty$ for $A\in\mathcal{A}$ in both the source and target domain. This metric also indicates the difference of data source attributes because for unseen incoming attributes, at least one entities in a pair should have the missing value. 
The result is depicted in Figure~\ref{fig:data_stats}. 
Ideally, the percentage bars included in this plot should all be close to 1, and this holds for both data in the source and target domain. 
In fact, we observe this pattern in the benchmark datasets~\cite{mudgal2018deep} (such as Beer, DBLP-ACM, etc.), which indicates few missing values and no significantly different attributes.
For the \texttt{Monitor} dataset, however, the pattern is different. 
We first observe that only 2 attributes (\ie, ``page\_title'' and ``source'') are close-to-1, while for all the remaining 11 attributes, less than 50\% entity pairs have complete attribute values. 
Such data sparsity reflects the challenge (\textbf{C1}).
In addition, we observe that the percentages of pairs without missing values are significantly different for the source and target domain. Particularly, we find that there are 5 out of 13 attributes only have non-missing entity pairs only in the target domain, which can be seen as new attributes (\textbf{C2}).
}

\begin{figure}[hb]
\captionsetup[subfigure]{justification=centering}
  \centering
  \begin{subfigure}[t]{0.245\textwidth}
      \centering
      \includegraphics[width=0.98\textwidth]{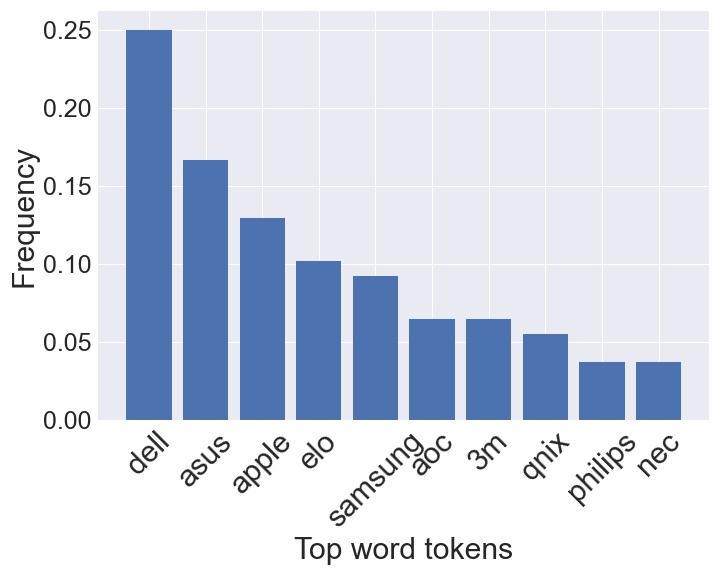}
          \caption{Frequency of top 10 word tokens in the source domain}
  	\label{fig:data_distri_a}
  \end{subfigure}
  ~
   \begin{subfigure}[t]{0.245\textwidth}
      \centering
      \includegraphics[width=0.98\textwidth]{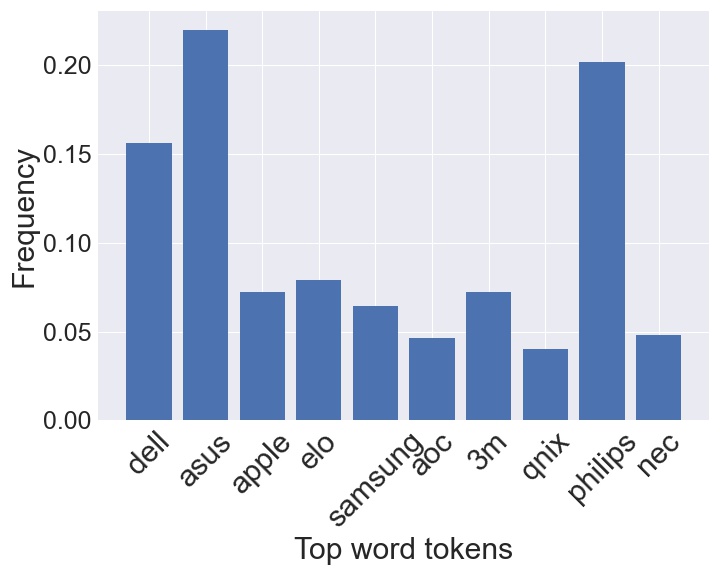}
          \caption{Frequency of top 10 word tokens in the target domain}
      \label{fig:data_distri_b}
  \end{subfigure}
  \vspace{-0.2cm}
  \caption{\texttt{Monitor}: the challenges of different attribute value distribution (\textbf{C3}) shown with the representative attribute ``prod\_type''. The frequency distribution of top 10 word tokens under this attribute is significantly different between the source and target domain. 
  }
\vspace{-.3cm}
\label{fig:data_distri}
\end{figure}

\dijin{
To illustrate the different attribute value distribution (\textbf{C3}), we showcase the attribute value distribution of one attribute, ``prod\_type'' as the representative. We plot the frequency distribution of 10 most frequently appearing word tokens. As shown in Figure~\ref{fig:data_distri}, the distributions of attribute ``prod\_type'' are quite different between the source and target domain, which indicates the challenge we attempt to address.
}

\clearpage
\thispagestyle{empty}

\subsection{Complete Experimental Results}
\label{sec:complete-result}

In this section we provide the complete numerical result shown in Section~\ref{sec:exp1}. 
Table~\ref{table:exp1-b} records the results on the \texttt{Monitor} dataset.
The results on the \texttt{Music} dataset are shown in Table~\ref{table:exp1-a}.

\begin{table}[hbp]
\centering
\setlength{\tabcolsep}{5pt} \caption{\method performance (PRAUC) on \texttt{Monitor}. All variants outperform the baseline, \methodC performs the best (marked in bold) with at least $0.51\%$ improvement over the second-best ($\ast$).}
\vspace{-0.3cm}
\resizebox{.96\columnwidth}{!}{
\begin{tabular}{@{}l|l|c|cH}
\toprule
 & Method & Overlapping & Disjoint & Track \\ \hline
\multicolumn{1}{c|}{\multirow{9}{*}{\texttt{Monitor}}}
& \dijin{TLER} & \dijin{0.4932 $\pm$ 0.0028 } & \dijin{0.3837 $\pm$ 0.0033 } \\
& DeepMatcher & 0.8336 $\pm$ 0.0032 &  0.7884 $\pm$ 0.0011 & x \\
& EntityMatcher & 0.8858 $\pm$ 0.0034 & 0.9051 $\pm$ $0.0042^{\ast}$ & x \\
& Ditto & 0.8841 $\pm$ 0.0010 & 0.8518 $\pm$ 0.0023 & x  \\
& CorDel-Attention & 0.7240 $\pm$ 0.0026 & 0.6353 $\pm$ 0.0165 & x  \\
\multicolumn{1}{c|}{} 
& \method-base & 0.8884 $\pm$ 0.0057 & 0.8711 $\pm$ 0.0050 & x  \\
\multicolumn{1}{c|}{} 
& \methodA & 0.8930 $\pm$ 0.0013 & 0.8719 $\pm$ 0.0030 & x  \\
\multicolumn{1}{c|}{} 
& \methodB & 0.9127 $\pm$ $0.0035^{\ast}$ & 0.9005 $\pm$ $0.0059$ & x  \\
\multicolumn{1}{c|}{} 
& \methodC & \textbf{0.9258 $\pm$ 0.0025} & \textbf{0.9106 $\pm$ 0.0029} & x  \\
\bottomrule
\end{tabular}
}
\label{table:exp1-b}
\end{table}

\begin{table*}[hb]
\centering
\vspace{-0.cm}
\setlength{\tabcolsep}{5pt} \caption{\method performance (PRAUC) of multi-source entity linkage on the \texttt{Music} data. The best score of each entity type is marked in bold. Out of \method variants, \methodC performs the best with $0.64\%\sim 5.50\%$ improvement over the second-best variant (marked with $\ast$) in PRAUC. \methodC outperforms the best-performing baselines (including \textsc{\method-base}) with $8.21\%$ improvement on average.
}
\vspace{-0.3cm}
\resizebox{0.96\linewidth}{!}{
\begin{tabular}{@{}l|l|ccc|ccc}
\toprule
 & \multirow{2}{*}{Method} & \multicolumn{3}{c|}{Overlapping ($\mathcal{D}^{\ast}_S\bigtimes \mathcal{D}^{\ast}_T$)} & \multicolumn{3}{c}{Disjoint ($\mathcal{D}^{\ast}_T\bigtimes \mathcal{D}^{\ast}_T$)}  \\ \cline{3-8}
 &  & Artist & Album & Track & Artist & Album & Track \\ \hline
\multicolumn{1}{c|}{\multirow{9}{*}{\texttt{Music-3K}}}
& \dijin{TLER} & \dijin{0.6454 $\pm$ 0.0021} & \dijin{0.5655 $\pm$ 0.0032} & \dijin{0.4263 $\pm$ 0.0011} & \dijin{0.4014 $\pm$ 0.0121} & \dijin{0.3605 $\pm$ 0.0033} & \dijin{0.4203 $\pm$ 0.0042} \\
& DeepMatcher & 0.6794 $\pm$ 0.0022 & 0.6093 $\pm$ 0.0009 & 0.5826 $\pm$ 0.0017 & 0.4492 $\pm$ 0.0021 & 0.3710 $\pm$ 0.0012 & 0.5572 $\pm$ 0.0014 \\
& EntityMatcher & 0.8682 $\pm$ 0.0017 & 0.6922 $\pm$ 0.0021 & 0.6694 $\pm$ 0.0084 & 0.6629 $\pm$ 0.0032 & 0.4733 $\pm$ 0.0014 & 0.6446 $\pm$ 0.0032 \\
& Ditto & 0.7920 $\pm$ 0.0032  & 0.6373 $\pm$ 0.0042  & 0.5938 $\pm$ 0.0051 & 0.5786 $\pm$ 0.0039 & 0.3832 $\pm$ 0.0027 & 0.5914 $\pm$ 0.0055  \\
& CorDel-Attention & 0.8489 $\pm$ 0.0047 & 0.6531 $\pm$ 0.0019 & 0.7032 $\pm$ 0.0364 & 0.7280 $\pm$ 0.0315 & 0.4586 $\pm$ 0.0002 & 0.6738 $\pm$ 0.0121 \\
& \method-base & 0.8545 $\pm$ 0.0143 & 0.7204 $\pm$ 0.0033 & 0.7277 $\pm$ 0.0077 & 0.7516 $\pm$ 0.0367 & 0.5569 $\pm$ 0.0072 & 0.7107 $\pm$ 0.0093 \\
\multicolumn{1}{c|}{} 
& \methodA & 0.9142 $\pm$ $0.0018^{\ast}$ & 0.7338 $\pm$ $0.0001^{\ast}$ & 0.7547 $\pm$ 0.0027 & 0.8263 $\pm$ $0.0121^{\ast}$ & 0.6071 $\pm$ $0.0072^{\ast}$ & 0.7453 $\pm$ 0.0012 \\
\multicolumn{1}{c|}{} 
& \methodB & 0.8633 $\pm$ 0.0011 & 0.7241 $\pm$ 0.0080 & 0.7904 $\pm$ $0.0048^{\ast}$ & 0.7510 $\pm$ 0.0331 & 0.5619 $\pm$ 0.0119 & 0.8129 $\pm$ $0.0057^{\ast}$ \\
\multicolumn{1}{c|}{} 
& \methodC & \textbf{0.9211 $\pm$ 0.0040} & \textbf{0.7833 $\pm$ 0.0031} & \textbf{0.8454 $\pm$ 0.0040} & \textbf{0.8390 $\pm$ 0.0125} & \textbf{0.6229 $\pm$ 0.0115} & \textbf{0.8193 $\pm$ 0.0097} \\\hline
\multicolumn{1}{c|}{\multirow{9}{*}{\texttt{Music-1M}}}
& \dijin{TLER} & \dijin{0.3384 $\pm$ 0.0013} & \dijin{0.2128 $\pm$ 0.0019} & \multicolumn{1}{c|}{\multirow{9}{*}{\texttt{--}}} & \dijin{0.2465 $\pm$ 0.0052} & \dijin{0.1237 $\pm$ 0.0031} & \multicolumn{1}{c}{\multirow{9}{*}{\texttt{--}}} \\
& DeepMatcher & 0.7132 $\pm$ 0.0033 & 0.5629 $\pm$ 0.0021 &  & 0.6033 $\pm$ 0.0045 & 0.1742 $\pm$ 0.0013 &  \\
& EntityMatcher & 0.8098 $\pm$ 0.0043 & 0.6731 $\pm$ 0.0024 & & 0.7239 $\pm$ 0.0038 & 0.2331 $\pm$ 0.0031 & \\
& Ditto & 0.7663 $\pm$ 0.0025 & 0.6123 $\pm$ 0.0022 &  & 0.6678 $\pm$ 0.0019 & 0.1933 $\pm$ 0.0027 &   \\
& CorDel-Attention & 0.8118 $\pm$ 0.0087 & 0.6811 $\pm$ 0.0432 &  & 0.7129 $\pm$ 0.0096 & 0.2224 $\pm$ 0.0010 &  \\
& \method-base & 0.8165 $\pm$ 0.0184 & 0.6872 $\pm$ 0.0053 &  & 0.7086 $\pm$ 0.0180 & 0.2269 $\pm$ 0.0050 &  \\
\multicolumn{1}{c|}{} 
& \methodA & 0.8607 $\pm$ $0.0066^{\ast}$ & 0.7693 $\pm$ $0.0038^{\ast}$ & & 0.7469 $\pm$ $0.0228^{\ast}$ & 0.3407 $\pm$ $0.0056^{\ast}$ &  \\
\multicolumn{1}{c|}{} 
& \methodB & 0.7942 $\pm$ 0.0090 & 0.7126 $\pm$ 0.0102 & & 0.7177 $\pm$ 0.0171 & 0.2473 $\pm$ 0.0131 &  \\
\multicolumn{1}{c|}{} 
& \methodC & \textbf{0.8710 $\pm$ 0.0130} & \textbf{0.7942 $\pm$ 0.0015} & & \textbf{0.7632 $\pm$ 0.0034} & \textbf{0.3582 $\pm$ 0.0043} &  \\
\bottomrule
\end{tabular}
}
\label{table:exp1-a}
\vspace{-0.2cm}
\end{table*}

\subsection{\methodC Algorithm}
\label{sec:algorithm-hyb}
Here we provide the detailed algorithm of \methodC, shown in Algorithm~\ref{algo:method-hybrid}.
\setlength{\textfloatsep}{8pt}
\begin{algorithm}[hbp]
\caption{\methodC}
\begin{algorithmic}[1]
\Ensure $\mathcal{D}_{S}=\{(\mathbf{h}_i, y_i)\}$, $\mathcal{S}_{U}=\{(\mathbf{h}_i, y_i)\}$, $\mathcal{D}_{T}=\{\mathbf{h}_i\}$, $\phi$, $B$

\Require Predicted $\hat{y_i}$ for $\mathbf{h}_i \in\mathcal{D}_{T}$, updated $\mathbf{a, W}$


\State Initialize $\mathbf{a, W}$ and $\mathbf{V, b}$

\Loop { training epochs}
    \For{$\mathbf{h}\in\mathcal{D}_S\cup\mathcal{D}_T\cup\mathcal{S}_U$} \label{alg3-line-affine-start}
        \State Form $\mathbf{x}$ with $\mathbf{V, b}$\Comment{Eq.~\eqref{eq:affine}}
    \EndFor \label{alg3-line-affine-end}
    
    \State $\bar{f}(\mathbf{x}^{\prime}) \leftarrow \frac{1}{|\mathcal{D}_{T}|}\sum_{\mathbf{x_i}\in\mathcal{D}_{T}}f(\mathbf{x_i})$  \label{alg3-line-target}

    \State $J \leftarrow 0$ \Comment{Initialize loss}
    \State $\mathcal{S}_{\text{batch}} \leftarrow \text{RANDOMSAMPLE}(\mathcal{D}_{S}, B)$


    \For{$(\mathbf{x},y) \in \mathcal{S}_{\text{batch}}$}  \label{alg1-line-attention-start}
        \State  $L_{\text{un}} \leftarrow (1-\lambda)L_{\text{base}} + \lambda L_{\text{target}}$ \Comment{Eq.~\eqref{eq:loss_un}}
        \State $J \leftarrow J + \nabla L_{\text{un}}$ 
        
    \EndFor \label{alg1-line-attention-end}

        
        
    \State Form $f$ with updated $\mathbf{a, W}$ \Comment{Eq.~\eqref{eq:fx}}
    \State Compute $\mathcal{D}_S^{+}$, $\mathcal{D}_S^{-}$, $\bar{d}^{+}_{\mathcal{D}_S}$, $\bar{d}^{-}_{\mathcal{D}_S}$  \Comment{Eq.~\eqref{eq:centroids}}
    
    \State $L_{\text{hybrid}} = L_{\text{un}} + \phi L_{\text{support}}$\Comment{Eq.~\eqref{eq:loss_hyb}}
    
    \State $J \leftarrow J + \nabla L_{\text{hybrid}}$ \Comment{Update $\mathbf{a}, \mathbf{W}, \mathbf{V}, \mathbf{b}$}

\EndLoop


    

\State Infer $\hat{\mathbf{y}}$ \Comment{Same as Line~\ref{alg1-line-reference-start-before}-~\ref{alg1-line-reference-end} of Algorithm~\ref{algo:method-un}}


\State \textbf{return} $\hat{\mathbf{y}}$, $\mathbf{a, W}$

\end{algorithmic}
\label{algo:method-hybrid}
\end{algorithm}

\end{document}